\documentclass[sigconf,natbib=false]{acmart}

\setcopyright{none}
\renewcommand\footnotetextcopyrightpermission[1]{}
\acmConference[2026]{}{}{}

\usepackage{tabularx}
\usepackage{balance}
\usepackage{enumitem}
\usepackage{multirow}
\usepackage[table]{xcolor}
\usepackage{algorithm}
\usepackage{algorithmic}
\usepackage{graphicx}
\usepackage{newfloat}
\usepackage{listings}
\DeclareCaptionStyle{ruled}{labelfont=normalfont,labelsep=colon,strut=off}
\floatstyle{ruled}
\newfloat{listing}{tb}{lst}{}
\floatname{listing}{Listing}

\AtBeginDocument{%
  }

\RequirePackage[
  datamodel=acmdatamodel,
  style=acmnumeric,
  ]{biblatex}

\begin{document}

\title{Multi-task Code LLMs: Data Mix or Model Merge?}

\author{Mingzhi Zhu}
\email{zhum8@rpi.edu}
\affiliation{%
  \institution{Rensselaer Polytechnic Institute}
  \city{Troy}
  \state{New York}
  \country{USA}
}

\author{Boris Sobolev}
\email{bsobolev@cisco.com}
\affiliation{%
  \institution{Cisco}
  \city{Woodbridge}
  \state{New Jersey}
  \country{USA}
}

\author{Rahul Krishna}
\email{rkrsn@ibm.com}
\affiliation{%
  \institution{IBM Research}
  \city{Yorktown Heights}
  \state{New York}
  \country{USA}
}

\author{Raju Pavuluri}
\email{pavuluri@us.ibm.com}
\affiliation{%
  \institution{IBM Research}
  \city{Yorktown Heights}
  \state{New York}
  \country{USA}
}
\author{Stacy Patterson}
\email{sep@cs.rpi.edu}
\affiliation{%
  \institution{Rensselaer Polytechnic Institute}
  \city{Troy}
  \state{New York}
  \country{USA}
}

\author{Michele Merler}
\email{mimerler@us.ibm.com}
\affiliation{%
  \institution{IBM Research}
  \city{Yorktown Heights}
  \state{New York}
  \country{USA}
}
\renewcommand{\shortauthors}{Zhu et al.}

\begin{abstract}
Recent research advocates deploying smaller, specialized code LLMs in agentic frameworks alongside frontier models, sparking interest in efficient strategies for multi-task learning that balance performance, constraints, and costs. We compare two approaches for creating small, multi-task code LLMs: data mixing versus model merging. We conduct extensive experiments across two model families (Qwen Coder and DeepSeek Coder) at two scales (2B and 7B parameters), fine-tuning them for code generation and code summarization tasks. Our evaluation on HumanEval, MBPP, and CodeXGlue benchmarks reveals that model merging achieves the best overall performance at larger scale across model families, retaining 96\% of specialized model performance on code generation tasks while maintaining summarization capabilities. Notably, merged models can even surpass individually fine-tuned models, with our best configuration of Qwen Coder 2.5 7B model achieving 92.7\% Pass@1 on HumanEval compared to 90.9\% for its task-specific fine-tuned equivalent. At a smaller scale we find instead data mixing to be a preferred strategy.
We further introduce a weight analysis technique to understand how different tasks affect model parameters and their implications for merging strategies. The results suggest that careful merging and mixing strategies can effectively combine task-specific capabilities without significant performance degradation, making them ideal for resource-constrained deployment scenarios. Our code can be found at \url{https://github.com/zmzfpc/Model_Merging_Data_Mixture}.
\end{abstract}






\maketitle

\section{Introduction}

\begin{figure}[t]
\centering
\includegraphics[width=\linewidth]{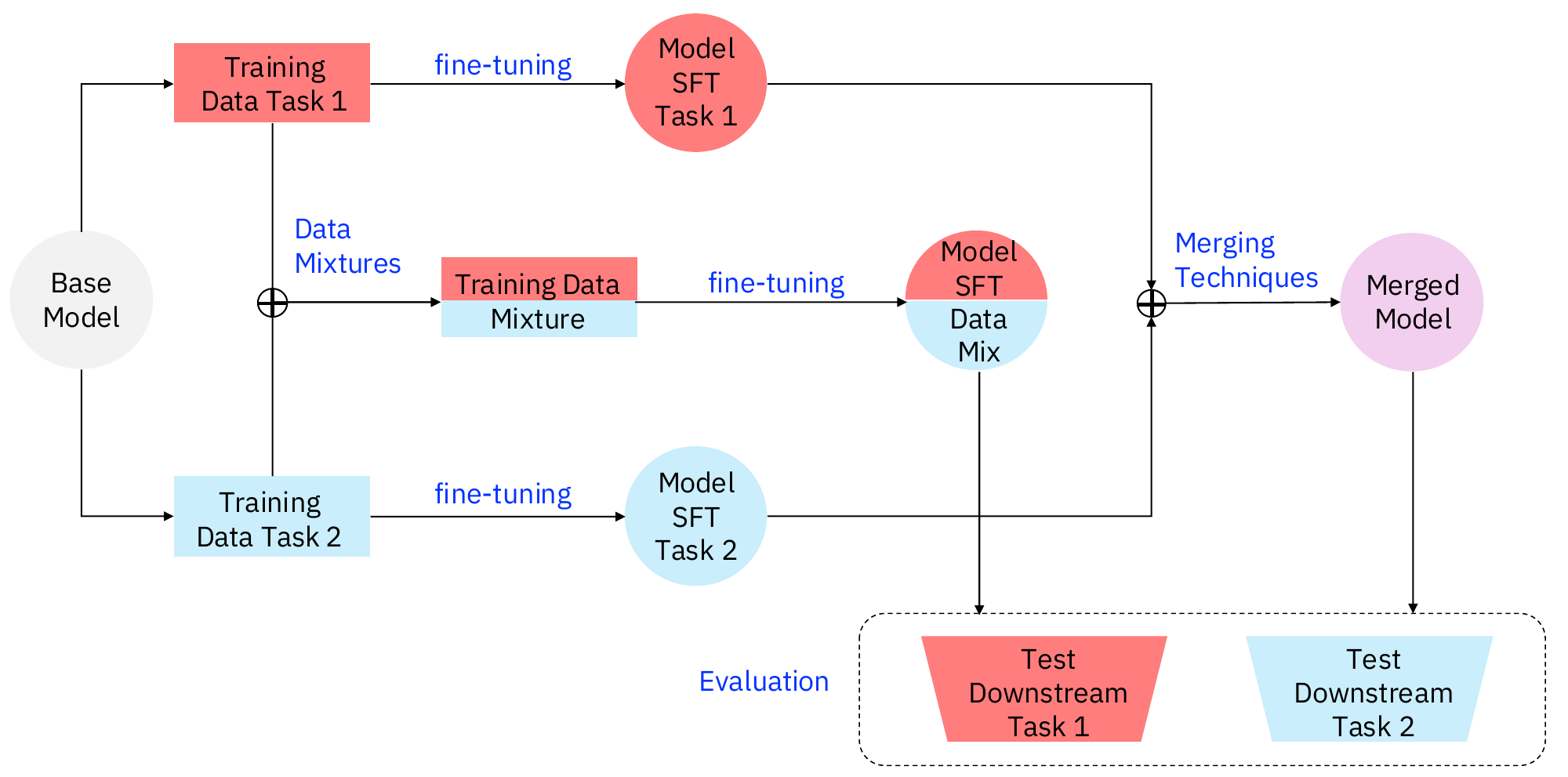}
\caption{Given a base code model and multiple downstream tasks, we investigate which strategy is optimal between model merging and data mixing to obtain a tuned multi-task model, based on model size and across model families.}
\label{fig:merge_diagram}
\end{figure}

Recently there has been more advocacy for smaller, specialized code LLMs, especially within the context of agentic frameworks~\cite{shen2024smallllmsweaktool}, as opposed to utilizing larger frontier models for every task ~\cite{belcak2025smalllanguagemodelsfuture}. Since they still need to handle more than one coding task in practice ~\cite{singhGundechaNeskovic2024CodeReviewSLM}, grouping such small specialized models into single ones that can equally perform on their specific tasks, or perhaps even benefit from related tasks, can be highly advantageous for resource savings. Striking the balance between hyper-specialization, resource constraints and effectiveness needs is dependent on multiple constraints in terms of cost, inference time, and quality of generations.

One way to obtain such performant multi-task small models is to mix the data for all desired tasks and data-mixture SFT (supervised fine-tune) once~\cite{lidata}. Another is to train separate specialists and merge their weights afterwards ~\cite{yang2024modelmergingllmsmllms}. 
While research is progressing substantially both in data-mixing ~\cite{raffel2020exploring,wei2022chain,shukor2025scalinglawsoptimaldata} as well as model merging ~\cite{yang2024modelmergingllmsmllms,yu2024languagemodelssupermario,zhou2025mergememodelmergingtechniques, Akiba_2025}, the comparative analysis and relationship between the two has been largely understudied for LLMs in general \cite{ahmadian2024mix}, and completely ignored for code LLMs specifically.

We aim to bridge that gap by investigating the following research questions, specifically tailored to code-specific applications:

\textit{RQ1} When building a single multi-task code LLM, is it more effective to (i) fine-tune once on a mixed multi-task dataset or (ii) fine-tune task specialists and merge their checkpoints?

\textit{RQ2} What role does scale, both in terms of number of parameters of a model and in tuning dataset size, play in answering RQ1?

\textit{RQ3} How do these strategies differ in the induced weight updates, and can weight-space signals predict which strategy will succeed?


To answer those questions, we make the following contributions:
\begin{itemize}[leftmargin=*]
 \item \textbf{Scale-swept comparison on code LLMs.} We study two code-LLM families at $\sim$2B and 7B parameters on two tasks (code generation and summarization), comparing data-mixture SFT against four strong merging methods.
    \item \textbf{Weight-space diagnostics for strategy selection.} Unlike prior work that treats weight deltas primarily as merging operators~\cite{li2025task,ilharco2022editing} or as coarse checkpoint-compatibility signals~\cite{hu2024exploring}, we use layer-wise weight shifts and inter-task update correlation as a mechanistic diagnostic that predicts whether data-mixture SFT or post-hoc merging will succeed for a given model scale.
    \item \textbf{Practical guidelines.} In our setting, $\sim$2B models consistently favor data-mixture SFT, while 7B models favor merging specialists, often with minimal or no average degradation relative to single-task checkpoints.
\end{itemize}


These insights can help practitioners select the right strategy for a given deployment budget, ensuring that small agentic systems can enjoy multi-task code intelligence without wasteful over-training.

\section{Related Work}

\subsection{LLM for Code}
Recently there has been an explosion of works focusing on training LLM architectures that were originally designed for NLP on coding, and specifically tuning them for tasks such as code generation~\cite{huynh2025codesurvey}, summarization~\cite{sun2024source}, test generation~\cite{pan2025asternaturalmultilanguageunit}, bug fixing~\cite{hossain2024deepdivelargelanguage}, vulnerability detection and repair~\cite{zhou2025large}, etc. Multiple surveys~\cite{zhang2024unifying, jiang2024survey, hou2024largelanguagemodelssoftware} on AI for code and LLM for software engineering cover this space. Several open source model families have been introduced including Llama~\cite{rozière2024codellamaopenfoundation}, Qwen~\cite{hui2024qwen2}, Deepseek~\cite{guo2024deepseek}, Granite~\cite{mishra2024granitecodemodelsfamily}, Gemma~\cite{codegemmateam2024codegemmaopencodemodels} and Devstral~\cite{devstral}, quickly closing the gap with frontier models.
These code LLMs have been powering the quickly proliferating field of AI coding assistants~\cite{sergeyuk2025using}.
A particularly active area of research is the development of coding agents \cite{jimenez2024swebenchlanguagemodelsresolve}, in which LLMs are provided the ability to interact with their environment, execute tool calls, observe the results of their actions and plan the next steps accordingly.
While frontier models have demonstrated remarkable capabilities for each of those tasks, there still remains a need for smaller, specialized models, both for agentic frameworks as well as for resource and budget constrained environments~\cite{shen2024smallllmsweaktool, belcak2025smalllanguagemodelsfuture}. Under those conditions, it is still desirable to use ``small" LLMs, and for multiple tasks. Good strategies to obtain such high-performing multi-task specialized models represent, however, a still unanswered question. 

\subsection{Model Merging}

Model merging has emerged as a highly efficient and cost-effective technique to combine multiple weights of checkpoints tuned on different tasks into a single model which retains as much as possible the performance of the originals on all the tasks. Several merging techniques have been proposed~\cite{yang2024modelmergingllmsmllms}, starting from basic linear weights combinations~\cite{wortsman2022modelsoupsaveragingweights} and spherical interpolation~\cite{jang2024sphericallinearinterpolationtextanchoring}, to task arithmetic on task vectors~\cite{yadav2023tiesmergingresolvinginterferencemerging} and random pruning and rescaling~\cite{yu2024languagemodelssupermario}. Methods have been developed also targeting specific architectures including mixture of experts~\cite{zhou2025mergememodelmergingtechniques, ding2024xftunlockingpowercode}, and specifically formalizing the objective of model merging into a multi-task learning framework~\cite{zhou-etal-2024-metagpt}. Recently, evolutionary techniques~\cite{Akiba_2025} have achieved state of the art performance across domains, including code generation. 
In the context of code repair, a ``continual merging'' paradigm has been proposed where task-specific adapters for code are sequentially merged to create a more powerful, multi-task model~\cite{dehghan2025mergerepairexploratorystudymerging}.
Some works have studied the performance of model merging techniques specifically for code generation across languages and tasks~\cite{Dixon1973270}, as well as for multiple diverse tasks, including coding~\cite{he2025mergebenchbenchmarkmergingdomainspecialized}.

\subsection{Multi-Task Fine-Tuning via Data Mixtures}

Multi-task supervised fine-tuning trains LLMs on combined datasets spanning multiple target tasks~\cite{raffel2020exploring,wei2022chain}. 
Code-related tasks including code generation and summarization are highly interrelated, suggesting that a joint model could share representations and even improve performance via cross-task knowledge transfer~\cite{liu2024mftcoder}. Thus, this approach can leverages shared knowledge across domains, where understanding code syntax benefits both generation and summarization tasks. 

Recent advances in data mixture optimization have established principled approaches for experimenting at smaller scales before transferring findings to larger models, following the hyperparameter transfer paradigm established in $\mu$P\cite{yang2022tensorprogramsvtuning}. This scaling methodology now extends to optimizing data mixtures, with emerging work on data mixing laws that predict language modeling performance~\cite{ye2025data} and scaling laws for optimal data combinations~\cite{shukor2025scalinglawsoptimaldata}. Additionally, research on critical mixture ratios for continual pre-training~\cite{gu2024cmr} and curriculum learning strategies~\cite{yoo2025codeswitching} demonstrates the growing complexity in data mixture design. These developments provide the foundation for systematic exploration of multi-task training strategies.

For small code LLMs, systematic evaluation is needed to determine whether merging task-specific fine-tuned models or fine-tuning on blended multi-task datasets is more effective. This work addresses the fundamental question of data mixing versus model merging for multi-task code LLMs.

\section{Experimental Setup}

\subsection{Model Families and Tasks} 
We conduct our experiments on two open-source code-focused LLM families, Qwen2.5-Coder~\cite{hui2024qwen2} and DeepSeekCoder~\cite{guo2024deepseek}, each at approximately 1.5B and 7B parameter scales. Models from these families have shown state of the art performance in their size cohorts at the time of their release on multiple coding tasks, making them strong candidates to specialize and analyze.
\begin{itemize}[leftmargin=*]
    \item \textbf{Qwen2.5-Coder} is a recent series of code-specialized models built on the Qwen2.5~\cite{team2024qwen2} architecture. We conducted experiments on the models from this family of 1.5B and 7B parameters.
    \item \textbf{DeepSeekCoder} is another modern code LM series trained on a large mixture of data including code (87\%) and text (13\%). We conducted experiments on the models from this family of 1.3B and 7B parameters.
\end{itemize}
For our study, we selected two representative yet sufficiently distinct coding tasks, which justifies the need for specialized fine-tuning for achieving optimal multi-task performance: 
\begin{enumerate}[leftmargin=*]
    \item \textbf{Code Generation}, i.e. writing code snippets given a natural language problem description. This is the most popular and studied code-related task. For evaluation of this task, we use the Humaneval~\cite{chen2021codex}, Humaneval+~\cite{evalplus}, MBPP~\cite{austin2021programsynthesislargelanguage}, and MBPP+~\cite{evalplus} benchmarks and report the pass@1 metric, that is, the proportion of prompts solved by the model with the first attempt.
    \item \textbf{Code Summarization}, i.e., generating a compact yet informative natural-language documentation for a snippet of given code. To evaluate code summarization, we use the test part~\cite{Zhang_2024} of the CodeXGLUE Code-to-Text dataset~\cite{lu2021codexglue} and measure performance using BLEU-4~\cite{papineni2002bleu}, chrF++~\cite{popovic2017chrf++}, ROUGE-L~\cite{lin2004rouge}, and METEOR~\cite{banerjee2005meteor} metrics, following common practice for measuring code generation/explanation quality in reference to golden, human written summaries~\cite{lu2021codexglue,vitale2025optimizingdatasetscodesummarization}.
\end{enumerate}

For both tasks, all our experiments were conducted for Python. We plan to explore other programming languages in future work.

\subsection{Supervised Fine-Tuning} 
We conducted supervised fine-tuning (SFT) experiments using two different strategies: (1) task-specific fine-tuning on individual tasks, (2) data mixture fine-tuning on combined datasets, and then merging these individually fine-tuned models. For code generation SFT, we leverage the KodCode dataset~\cite{xu2025kodcode}, a large-scale synthetic corpus of 268K coding problems with verified solutions and comprehensive test cases, to train models for accurate code synthesis. For code summarization fine-tuning, we utilize code-comment pairs from the CodeXGLUE Code-to-Text dataset~\cite{lu2021codexglue} training split containing 417K high-quality pairs of code snippets and their corresponding natural language descriptions. 
All fine-tuning was implemented using LLamaFactory~\cite{zheng2024llamafactory}, a unified framework for efficient large language model training. We conducted all experiments on 4$\times$H100 GPUs, with training times of approximately 4 hours for 1.5B models and 20 hours for 7B models. The training configuration and dataset statistics are detailed in Tables~\ref{tab:hyperparameters} and~\ref{tab:datasets}, respectively.

\paragraph{Data mixture.} When experimenting with mixing data, we simply combined the KodCode and CodeXGLUE datasets and shuffled them. For SFT on this mixed dataset, we utilized the same training hyperparameters as the task-specific SFT runs reported in Table~\ref{tab:hyperparameters}.

\begin{table}[t]
\centering
\caption{SFT Hyperparameters by Model Size}
\label{tab:hyperparameters}
\resizebox{\linewidth}{!}{%
\begin{tabular}{lcc}
\hline
\textbf{Hyperparameter} & \textbf{1.5B Models} & \textbf{7B Models} \\
\hline
Learning Rate & 1e-5 & 5e-6 \\
Batch Size(effective) & 128 & 64 \\
Gradient Accumulation Steps & 8 & 8 \\
Max Sequence Length & 4,096 & 4,096 \\
Training Epochs & 2 & 2 \\
Warmup Ratio & 0.03 & 0.03 \\
LR Scheduler & Cosine & Cosine \\
Optimizer & AdamW & AdamW \\
\hline
\end{tabular}
}
\end{table}

\begin{table}[t]
\centering
\caption{Training and Evaluation Dataset Statistics}
\label{tab:datasets}
\resizebox{\linewidth}{!}{%
\begin{tabular}{lccc}
\hline
\textbf{Dataset} & \textbf{Task} & \textbf{Train Split} & \textbf{Test Split} \\
\hline
CodeXGLUE  & Summarization & 417K & 25,834\\
KodCode & Generation & 268K & - \\
HumanEval & Generation & - & 164 \\
MBPP & Generation & - & 379 \\
\hline
\end{tabular}
}
\end{table} 

\subsection{Model Merging Methods} 
To merge the specialized models, we employ MergeKit \cite{goddard2025arceesmergekittoolkitmerging}, a comprehensive toolkit for combining multiple fine-tuned models. We evaluate four state-of-the-art merging strategies: 
\begin{itemize}[leftmargin=*]
    \item \textbf{Linear Merge} \cite{wortsman2022modelsoupsaveragingweights}, performs simple parameter averaging between models
    \item \textbf{DARE} \cite{yu2024languagemodelssupermario}, selectively drops and rescales parameters reducing interference between task-specific knowledge
    \item \textbf{TIES} \cite{yadav2023tiesmergingresolvinginterferencemerging} (Trim, Elect Sign \& Merge), resolves sign conflicts by trimming redundant parameters and electing consistent signs before merging
    \item \textbf{DELLA} \cite{deep2024dellamergingreducinginterferencemodel} (Deep Linear Learning Alignment), reduces interference through aligned parameter updates
\end{itemize}
 Each method represents a different approach to handling the challenge of preserving task-specific capabilities while minimizing negative interference when combining specialized models.

\section{Model Weights Analysis}

\subsection{Data Mixtures vs. Model Merging} 

We present a comprehensive analysis of how different fine-tuning strategies affect transformer layers across model scales. In order to do so, we measured across four architectures (Qwen-1.5B/7B and DeepSeek-1.3B/7B) per-layer weight shifts using average $L_2$ distance between the SFT and the base models.

\begin{figure*}[ht]
    \centering
    \includegraphics[width=\linewidth]{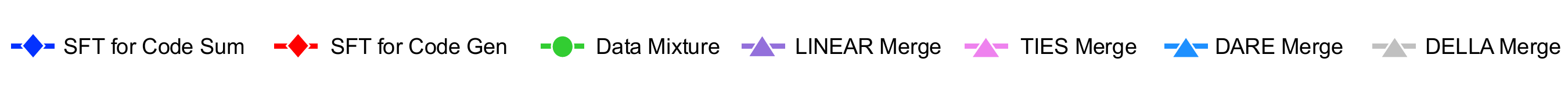}
    \begin{tabular}{cc}
    \includegraphics[width=\columnwidth]{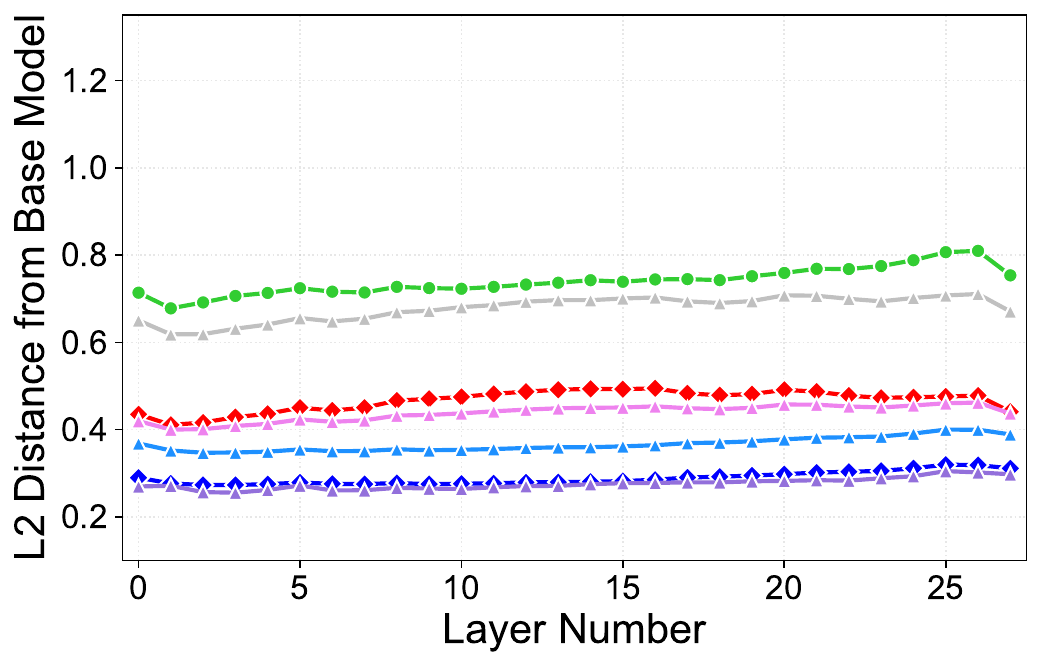} &
     \includegraphics[width=\columnwidth]{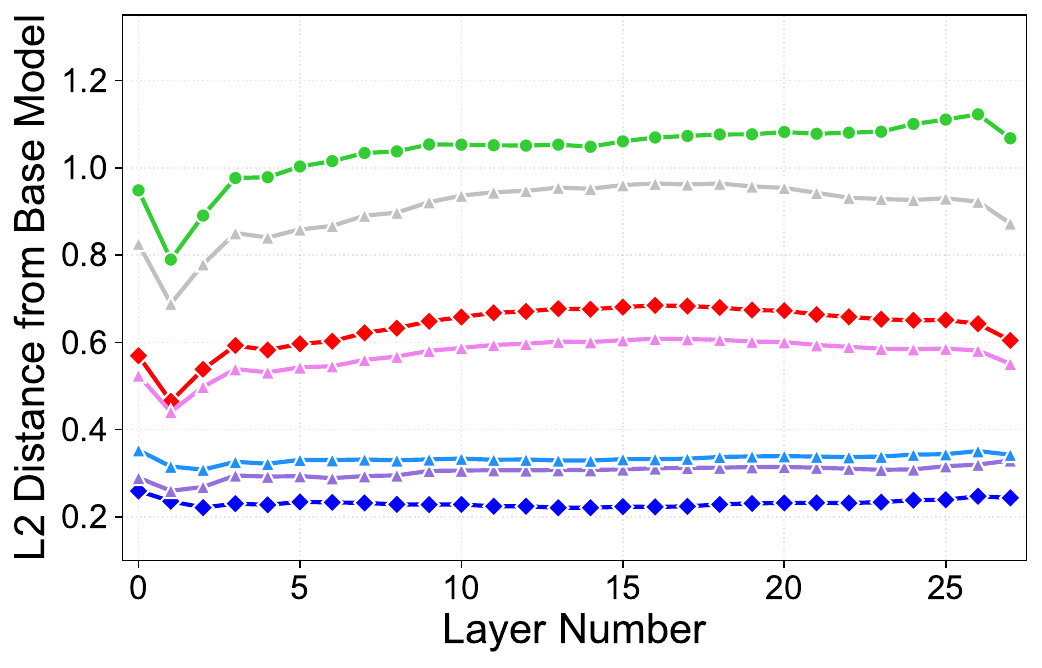}\\
    (a)  Qwen 1.5B & (b) Qwen 7B\\
    \includegraphics[width=\columnwidth]{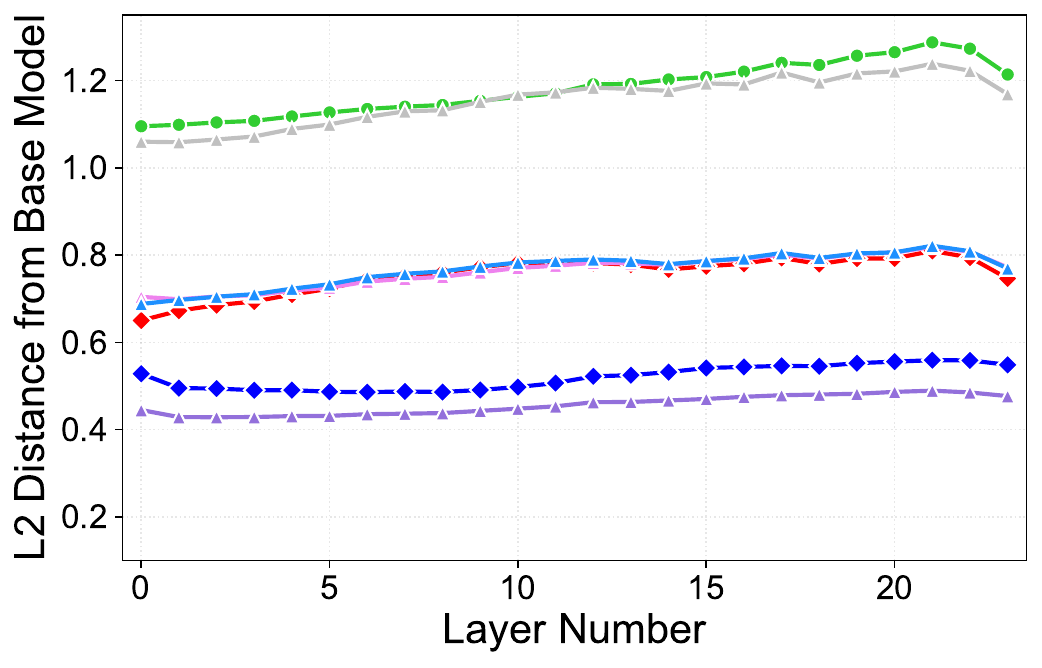}  & 
    \includegraphics[width=\columnwidth]{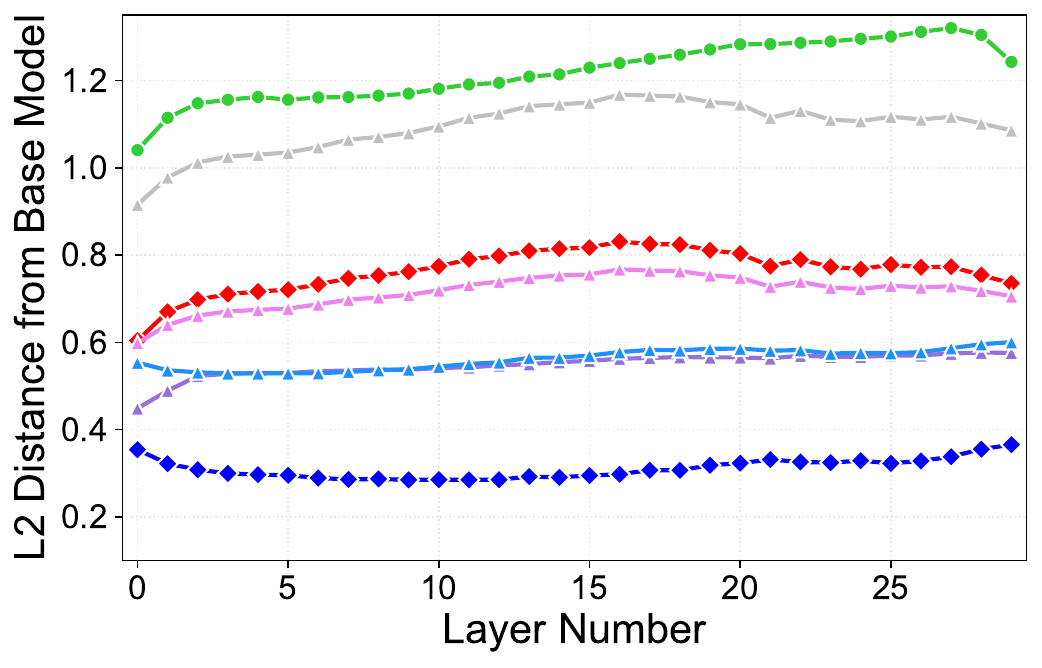}\\   
    (c) DeepSeek 1.3B & (d) DeepSeek 7B
     \end{tabular}
    \caption{Layerwise L$_2$ distance from the base model for each fine-tuned or merged variant, across four model architectures. Each subplot corresponds to one base model and parameter scale: (a)~Qwen~1.5B, (b)~Qwen~7B, (c)~DeepSeek~1.3B, and (d)~DeepSeek~7B. Within each subplot, curves are shown for the two single-task fine-tunes (CodeGen and CodeSum), the multi-task data-mix fine-tune, and the four merged models (Linear, TIES, DARE, DELLA). Lower values indicate that the post-trained model's weights remain closer to the original base model at that layer.}
    \label{fig:l2-distance}
\end{figure*}

Figure~\ref{fig:l2-distance} (a) to (d) shows the layer-wise L$_2$ norm of weight differences with respect to the base models for all seven variants: the two single-task SFT models, the multi-task data mixture SFT models, and the four models merged with Linear, TIES, DARE and DELLA. 

Across all four model architectures, the Data Mixture approach with green triangular line consistently exhibits the highest $L_2$ norms throughout the network layers, ranging from approximately 0.7--1.3 depending on model configuration. At parity of hyperparameters, more diverse and unique training examples translate to larger weights shifts. In contrast, model merging methods demonstrate substantially lower $L_2$ norms, falling between 0.3--0.8, while individual SFT models show the most conservative parameter modifications. Specifically, the individual SFT model for Code Summarization shows the lowest $L_2$ norms (especially at larger scales), suggesting minimal parameter deviation from the base model. The Code Generation SFT model shows moderate $L_2$ norms, as it modifies a few layers more significantly than the Code Summarization model, but still remains below the Data Mixture SFT.

The difference in $L_2$ norms between the data mixture and model merging approaches reveals fundamental distinctions in how these methods combine task-specific knowledge. Data mixture, which trains on combined datasets from multiple tasks simultaneously, induces larger parameter changes across all layers, as evidenced by its consistently elevated $L_2$ norms. This suggests that the model must learn more complex parameter configurations to accommodate the diverse training signals present in the mixed data. The high magnitude of these changes indicate that data mixture forces the model to find a compromise representation that can handle multiple tasks, potentially leading to interference effects where task-specific optimizations conflict with each other.

When comparing merging strategies, we find that they generally produce more conservative weight updates than direct fine-tuning, but with some differences in how they affect the network. Linear merging
reduces extreme weight shifts to some extent; its L$_2$ curve lies below the single-task curves at the peak layers. TIES  merging heuristically resolves conflicting weight updates, resulting in L$_2$ distances that closely track the union of the two task-specific profiles: in layers that one or both tasks significantly adjusted, TIES retains those large changes, whereas in layers with smaller or opposing task updates, it prunes or cancels them. DARE and DELLA yield the most conservative changes instead. By dropping many minor weight updates and rescaling the remaining ones, these methods keep the majority of layers very close to the original weights. Their L$_2$ curves stay low and flat for most layers, with only a few layers showing any substantial deviation from the base model. In effect, DARE/DELLA concentrate the merged adaptation into a limited set of significant weight shifts, minimizing alterations elsewhere.

Interestingly, while the majority of model merging techniques produce as expected weight shifts that are contained between the two task-specific SFT models they are combining, DELLA deviates significantly, almost reaching the same levels of variations from the base model as the ones introduced by data mixing. 

We observe that Deepseek models undergo larger weight shifts with respect to their Qwen counterparts at the same scale. Looking at the results in the Table \ref{tab:results}, we hypothesize that such larger shifts were needed to more aggressively adapt to the downstream tasks since the performance of the base models is lower for Deepseek models compared to Qwen. We can see that the larger weight shifts in Deepseek models correspond to a larger delta of relative improvement on average of the task specific SFT models with respect to the base ones (19.11\% and 17.79\% for 1.3B and 7B respectively, as opposed to 13.47\% and 12.8\% for their Qwen counterparts).  

\subsection{Pearson Correlation Analysis} 
\begin{figure*}[ht]
    \centering
    \begin{tabular}{cc}
    \includegraphics[width=\columnwidth]{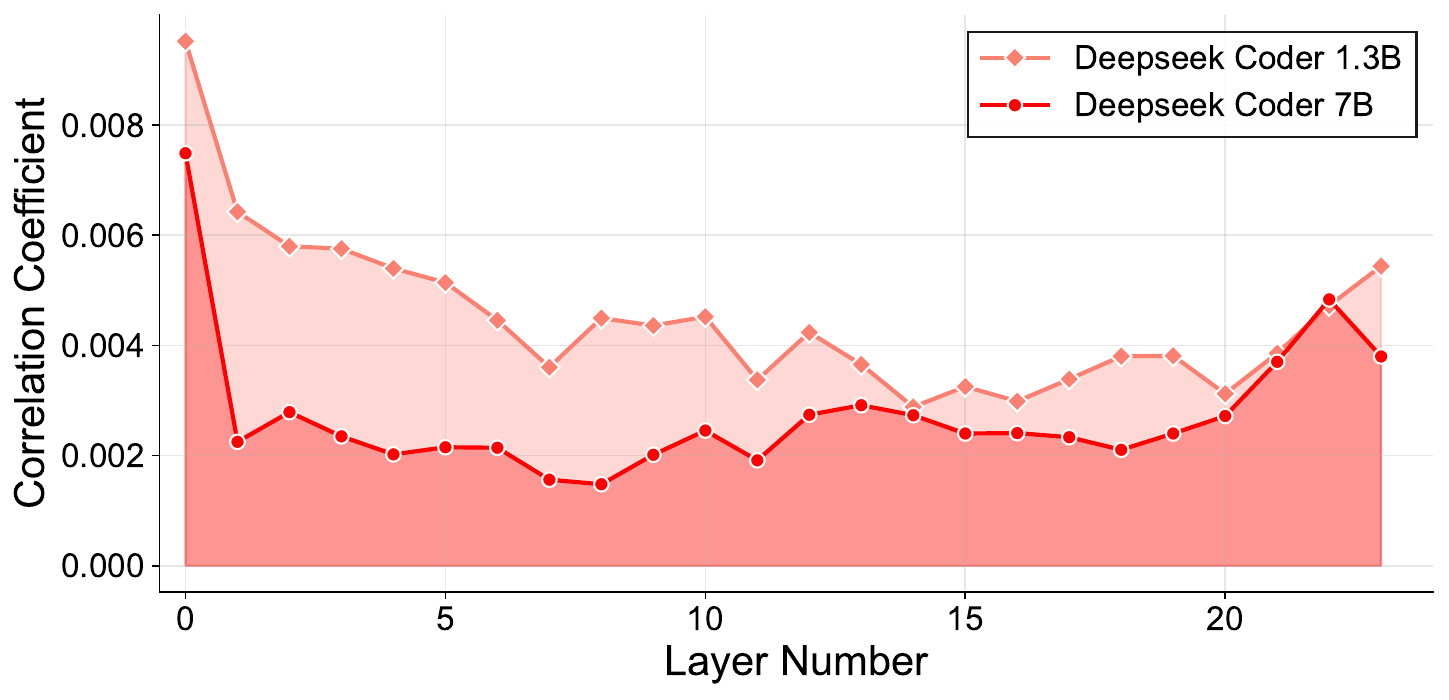}
         &  \includegraphics[width=\columnwidth]{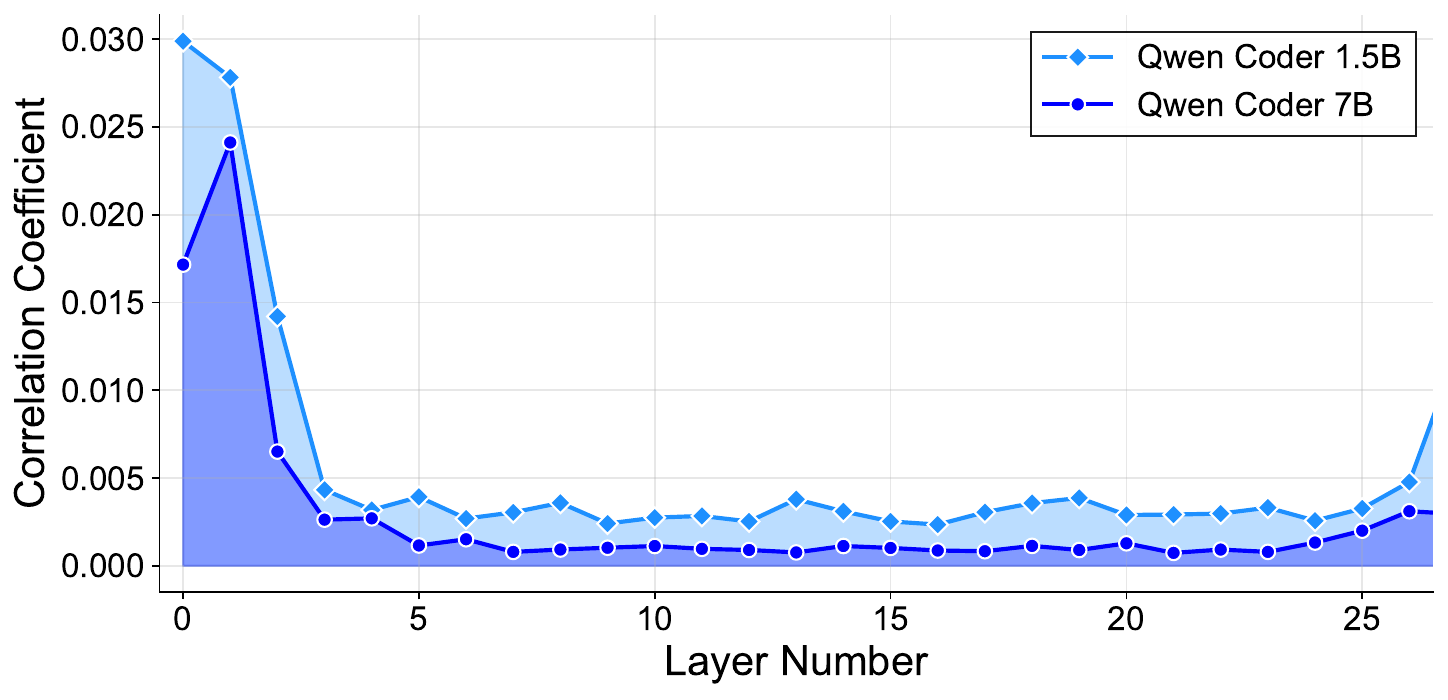}\\
        (a) & (b)
    \end{tabular}  
    \vspace{-0.2cm}
    \caption{Layerwise Pearson correlation between the weight updates of code-generation vs.\ code-summarization fine-tuned models for (a) DeepSeek and (b) Qwen families. Smaller models (Qwen~1.5B, DeepSeek~1.3B) show consistently higher correlation coefficients across layers, whereas larger models (7B) exhibit lower correlation.}
    \label{fig:correlation}
\end{figure*}

To understand why certain models favor merging while others prefer data mixing, we utilize the Pearson correlation~\cite{benesty2009pearson} $r_{\Delta w} $ between task-specific weight updates:
\begin{equation}
r_{\Delta w_j} = \frac{\sum_{i=1}^{n}(\Delta w_{ij}^{g} - \overline{\Delta w^{g}_j})(\Delta w_{ij}^{s} - \overline{\Delta w^{s}_j})}{\sqrt{\sum_{i=1}^{n}(\Delta w_{ij}^{g} - \overline{\Delta w^{g}_j})^2}\sqrt{\sum_{i=1}^{n}(\Delta w_{ij}^{s} - \overline{\Delta w^{s}_j})^2}}
\end{equation}
where $\Delta w_{ij}^{g}$ and $\Delta w_{ij}^{s}$ represent the weight changes for code generation and summarization tasks at parameter $i$ of layer $j$, respectively. $\overline{\Delta w^{g}_j}$ and $\overline{\Delta w^{s}_j}$ represent the mean weight changes for each task at layer $j$. Pearson correlation measures the linear relationship between two variables, ranging from -1 (perfect negative correlation) to +1 (perfect positive correlation). High correlation between task-specific weight changes indicates overlapping adaptations, which can lead to interference during merging. Conversely, low correlation suggests orthogonal task representations that can be effectively combined without conflict.
We compute layer-wise Pearson correlation between weight changes ($\Delta w$) of code generation and summarization models relative to their shared base model. Figure~\ref{fig:correlation} reveals a clear scale-dependent pattern: smaller models (Qwen~1.5B, DeepSeek~1.3B) exhibit consistently high correlation across layers, while larger models (7B variants) show substantially lower correlation. Thus small models adapt to both tasks by modifying similar weight subsets, creating interference during merging. Large models utilize distinct representational pathways for each task, enabling successful weight-space combination. These findings correlate with our empirical results: high correlation models benefit from data mixing to avoid parameter conflicts, while low correlation models excel at model merging due to orthogonal task representations.

\begin{table*}[t!] 
\centering
\caption{Performance of Merged vs. Fine-Tuned Models on Code Generation and Summarization. We report code generation success (pass@1 on HumanEval and MBPP) and code summarization quality (BLEU-4, chrF++, ROUGE-L, METEOR) for Qwen2.5-Coder and DeepSeekCoder (1.5B and 7B). Pretrained = zero-shot base model; SFT-CodeGen = fine-tuned on code generation; SFT-CodeSum = fine-tuned on summarization; Data-Mixture = fine-tuned on combined data; Linear/TIES/DARE/DELLA = merged models using those methods.}
\label{tab:results}
\resizebox{\textwidth}{!}{
\begin{tabular}{l|cccc|cccc|c}
\hline
Task & \multicolumn{4}{c|}{\textbf{Code Generation}} &  \multicolumn{4}{c|}{\textbf{Code Summarization}} &\\ 
Dataset  & HumanEval & HumanEval+ & MBPP & MBPP+ & \multicolumn{4}{c|}{CodeXGLUE Code-to-text} &  \\
Metric & pass@1 $\uparrow$& pass@1 $\uparrow$  & pass@1 $\uparrow$  & pass@1 $\uparrow$  & BLEU-4$\uparrow$ & chrF++$\uparrow$& ROUGE-L$\uparrow$ & METEOR $\uparrow$& AVG \% wrt SFT  \\
\hline
\multicolumn{10}{c}{} \\
\multicolumn{1}{c}{} & \multicolumn{9}{c}{\textit{Qwen2.5-Coder 1.5B}} \\
\hline
Base Model & 0.701(-7.28\%) & 0.640(-9.48\%) & 0.690(-9.09\%) & 0.590(-10.06\%) & 0.0259(-47.68\%) & 0.3144(-2.15\%) & 0.2638(-19.55\%) & 0.3015(-2.46\%) & -13.47\% \\
SFT-CodeGen/Sum  & 0.756(0\%) & 0.707(0\%) & 0.759(0\%) & 0.656(0\%) & 0.0495(0\%) & 0.3213(0\%) & 0.3279(0\%) & 0.3091(0\%) & 0.00\% \\
\rowcolor{lightgray} Data-Mixture SFT & \textbf{0.768(+1.59\%)} & \textbf{0.707(0\%)} & \textbf{0.754(-0.66\%)} & \textbf{0.646(-1.52\%)} & \textbf{0.0477(-3.64\%)} & \textbf{0.3148(-2.02\%)} & \textbf{0.3216(-1.92\%)} & \textbf{0.3002(-2.88\%)} & \textbf{-1.26\%} \\
Linear Merge & 0.671(-11.24\%) & 0.598(-15.42\%) & 0.749(-1.32\%) & 0.651(-0.76\%) & 0.0434(-12.32\%) & 0.306(-4.76\%) & 0.3086(-5.89\%) & 0.2912(-5.79\%) & -7.19\% \\
TIES Merge & 0.652(-13.76\%) & 0.591(-16.41\%) & 0.754(-0.66\%) & 0.635(-3.20\%) & 0.0372(-24.85\%) & 0.3024(-5.88\%) & 0.2894(-11.74\%) & 0.2883(-6.73\%) & -10.40\% \\
DARE Merge & 0.616(-18.52\%) & 0.567(-19.80\%) & 0.725(-4.48\%) & 0.638(-2.74\%) & 0.0468(-5.45\%) & 0.3157(-1.74\%) & 0.3205(-2.26\%) & 0.3033(-1.88\%) & -7.11\% \\
DELLA Merge & 0.573(-24.21\%) & 0.524(-25.88\%) & 0.712(-6.19\%) & 0.622(-5.18\%) & 0.0183(-63.03\%) & 0.2422(-24.62\%) & 0.2041(-37.76\%) & 0.2033(-34.23\%) & -27.64\% \\
\hline
\multicolumn{10}{c}{} \\
\multicolumn{1}{c}{}  & \multicolumn{9}{c}{\textit{Qwen2.5-Coder 7B}} \\
\hline
Base Model & 0.880(-3.19\%) & 0.835(-3.58\%) & 0.825(-6.04\%) & 0.717(-3.50\%) & 0.0284(-55.28\%) & 0.3190(-4.95\%) & 0.2742(-20.43\%) & 0.3081(-5.46\%) & -12.80\% \\
SFT-CodeGen/Sum & 0.909(0\%) & 0.866(0\%) & 0.878(0\%) & 0.743(0\%) & 0.0635(0\%) & 0.3356(0\%) & 0.3446(0\%) & 0.3259(0\%) & 0.00\% \\
Data-Mixture SFT & 0.902(-0.77\%) & 0.829(-4.27\%) & 0.857(-2.39\%) & 0.733(-1.35\%) & 0.0598(-5.83\%) & 0.3303(-1.58\%) & 0.3367(-2.29\%) & 0.3191(-2.09\%) & -2.57\% \\
Linear Merge & 0.835(-8.14\%) & 0.787(-9.12\%) & 0.841(-4.21\%) & 0.712(-4.17\%) & 0.0601(-5.35\%) & 0.3298(-1.73\%) & 0.3419(-0.78\%) & 0.3175(-2.58\%) & -4.51\% \\
TIES Merge & 0.866(-4.73\%) & 0.811(-6.35\%) & 0.862(-1.82\%) & 0.728(-2.02\%) & 0.0593(-6.61\%) & 0.3333(-0.69\%) & 0.3428(-0.52\%) & 0.3215(-1.35\%) & -3.01\% \\
DARE Merge & 0.829(-8.80\%) & 0.780(-9.93\%) & 0.815(-7.18\%) & 0.704(-5.25\%) & 0.0614(-3.31\%) & 0.3340(-0.48\%) & 0.3459(+0.38\%) & 0.3235(-0.74\%) & -4.41\% \\
\rowcolor{lightgray} DELLA Merge & \textbf{0.902(-0.77\%)} & \textbf{0.841(-2.89\%)} & \textbf{0.849(-3.30\%)} & \textbf{0.735(-1.08\%)} & \textbf{0.0578(-8.98\%)} & \textbf{0.3398(+1.25\%)} & \textbf{0.3391(-1.60\%)} & \textbf{0.3284(+0.77\%)} & \textbf{-2.08\%} \\
\hline
\multicolumn{10}{c}{} \\
\multicolumn{1}{c}{}  & \multicolumn{9}{c}{\textit{DeepSeek-Coder 1.3B}} \\
\hline
Base Model & 0.634(-10.33\%) & 0.598(-10.88\%) & 0.608(-3.49\%) & 0.516(-5.84\%) & 0.0181(-61.49\%) & 0.2786(-13.26\%) & 0.2157(-33.49\%) & 0.2653(-14.09\%) & -19.11\% \\
SFT-CodeGen/Sum & 0.707(0\%) & 0.671(0\%) & 0.630(0\%) & 0.548(0\%) & 0.0470(0\%) & 0.3212( 0\%) & 0.3243(0\%) & 0.3088(0\%) & 0.00\% \\
\rowcolor{lightgray} Data-Mixture SFT & \textbf{0.671(-5.09\%)} & \textbf{0.634(-5.51\%)} & \textbf{0.647(+2.70\%)} & \textbf{0.556(+1.46\%)} & \textbf{0.0434(-7.66\%)} & \textbf{0.3212(0\%)} & \textbf{0.3212(-0.96\%)} & \textbf{0.3087(-0.03\%)} & \textbf{-1.89\%} \\
Linear Merge & 0.628(-11.17\%) & 0.561(-16.39\%) & 0.646(+2.54\%) & 0.558(+1.82\%) & 0.0460(-2.13\%) & 0.3309(+3.02\%) & 0.3127(-3.58\%) & 0.3208(+3.89\%) & -2.75\% \\
TIES Merge & 0.610(-13.72\%) & 0.561(-16.39\%) & 0.622(-1.27\%) & 0.545(-0.55\%) & 0.0452(-3.83\%) & 0.3213(+0.03\%) & 0.3007(-7.28\%) & 0.3129(+1.33\%) & -5.21\% \\
DARE Merge & 0.616(-12.87\%) & 0.561(-16.39\%) & 0.640(+1.59\%) & 0.553(+0.91\%) & 0.0454(-3.40\%) & 0.3296(+2.62\%) & 0.3102(-4.35\%) & 0.3200(+3.63\%) & -3.53\% \\
DELLA Merge & 0.555(-21.50\%) & 0.506(-24.59\%) & 0.563(-10.63\%) & 0.489(-10.77\%) & 0.0403(-14.26\%) & 0.2978(-7.29\%) & 0.2743(-15.42\%) & 0.2909(-5.80\%) & -13.78\% \\
\hline
\multicolumn{10}{c}{} \\
\multicolumn{1}{c}{}  & \multicolumn{9}{c}{\textit{DeepSeek-Coder 7B}} \\
\hline
Base Model & 0.713(-10.09\%) & 0.640(-14.67\%) & 0.754(-3.33\%) & 0.664(-1.46\%) & 0.0203(-70.00\%) & 0.3063(-6.56\%) & 0.2243(-33.32\%) & 0.3081(-2.87\%) & -17.79\% \\
SFT-CodeGen/Sum  & 0.793(0\%) & 0.750(0\%) & 0.775(0\%) & 0.672(0\%) & 0.0530(0\%) & 0.3288(0\%) & 0.3362(0\%) & 0.3172(0\%) & 0.00\% \\
Data-Mixture SFT & 0.762(-3.91\%) & 0.720(-4.00\%) & 0.786(+1.75\%) & 0.680(+1.46\%) & 0.0540(+1.49\%) & 0.3311(+0.95\%) & 0.3361(-0.03\%) & 0.3200(+0.95\%) & -0.17\% \\
Linear Merge & 0.738(-6.94\%) & 0.689(-8.13\%) & 0.775(0\%) & 0.675(+0.55\%) & 0.0512(-4.26\%) & 0.3326(+1.46\%) & 0.3276(-2.62\%) & 0.3209(+1.17\%) & -2.35\% \\
TIES Merge & 0.732(-7.69\%) & 0.689(-8.13\%) & 0.786(+1.75\%) & 0.688(+2.92\%) & 0.0509(-4.89\%) & 0.3332(+1.65\%) & 0.3178(-5.53\%) & 0.3228(+1.77\%) & -2.27\% \\
\rowcolor{lightgray} DARE Merge & \textbf{0.762(-3.91\%)} & \textbf{0.734(-2.13\%)} & \textbf{0.799(+3.81\%)} & \textbf{0.701(+5.29\%)} & \textbf{0.0531(+0.21\%)} & \textbf{0.3315(+1.13\%)} & \textbf{0.3203(-4.79\%)} & \textbf{0.3196(+0.76\%)} & \textbf{-0.01\%} \\
DELLA Merge & 0.720(-9.21\%) & 0.671(-10.53\%) & 0.778(+0.48\%) & 0.669(-0.55\%) & 0.0515(-3.62\%) & 0.3305(+0.82\%) & 0.3117(-7.34\%) & 0.3202(+0.95\%) & -3.63\% \\
\hline
\end{tabular}
}
\end{table*}

\section{Benchmark Experiment Results Analysis}\label{sec:results}
Table~\ref{tab:results} presents the performance comparison between merged and fine-tuned models across code generation and summarization tasks. The results reveal a relationship between model scale and optimal multi-task learning strategy: smaller models consistently achieve superior performance through data mixture approaches, while larger models benefit more from post-hoc model merging.

\subsection{Scale-Dependent Strategy Effectiveness}
The experimental results demonstrate that model capacity fundamentally determines whether joint training or weight-space merging yields better multi-task performance. 

For the Qwen2.5-Coder 1.5B model, data-mixture SFT achieves results close to task-specific specialists with only -1.26\% average performance degradation, even improving HumanEval pass@1 from 0.756 to 0.768. In contrast, model merging approaches suffer significant performance drops, with Linear and DARE merging incurring approximately 7\% average loss, while TIES and DELLA degrade by -10.40\% and -27.64\% respectively. This pattern suggests that smaller models lack sufficient capacity to accommodate conflicting task-specific weight updates without destructive interference.

The DeepSeek-Coder 1.3B model exhibits similar behavior, where data-mixture SFT maintains competitive performance with -1.89\% average loss while significantly outperforming the best merging approach. Interestingly, Linear merging shows selective improvements on text metrics, chrF++ increases from 0.3212 to 0.3309, but at the cost of substantial code generation capability, approximately 11\% degradation on HumanEval. This trade-off indicates that merging algorithms struggle to preserve both capabilities simultaneously in capacity-constrained models.

\begin{table*}[t!]
\centering
\caption{Effect of training dataset size on specialist SFT performance.  
Code generation quality is measured by HumanEval+\cite{evalplus} pass@1; code summarization quality by BLEU-4\cite{papineni2002bleu} on CodeXGLUE test set\cite{lu2021codexglue}.}
\label{tab:rq2_scale}
\begin{tabular}{l|ccc|ccc}
\hline
\textbf{Model} &
\multicolumn{3}{c|}{\textbf{Code Generation}\,(pass@1)\,$\uparrow$} &
\multicolumn{3}{c}{\textbf{Code Summarization}\,(BLEU-4)\,$\uparrow$} \\
\hline
Sub-sampling Ratio & 25\% & 50\% & 100\% & 25\% & 50\% & 100\% \\
\hline
Qwen2.5-Coder 1.5B & 0.671  & 0.701 & 0.707 & 0.0303 & 0.0317 & 0.0495 \\
Qwen2.5-Coder 7B   & 0.841  & 0.854  & 0.866 & 0.0351 & 0.0428 & 0.0635 \\
DeepSeek-Coder 1.3B & 0.610  & 0.656  & 0.671 & 0.0215 & 0.0304 & 0.0470 \\
DeepSeek-Coder 7B   & 0.720 & 0.746  & 0.750 & 0.0299 & 0.0349 & 0.0530 \\
\hline
\end{tabular}
\end{table*}
\subsection{Emergence of Merging Advantages at Scale}
The effectiveness pattern reverses dramatically for 7B-parameter models. In Qwen2.5-Coder 7B, DELLA merging achieves the best overall performance with only -2.08\% average loss, outperforming data-mixture SFT with -2.57\%. The merged model maintains strong code generation performance, HumanEval pass@1 of 0.902 versus 0.909 for specialists, while effectively preserving summarization capabilities. This suggests that larger models can accommodate multiple task-specific adaptations within the same parameter space without significant interference.

DeepSeek-Coder 7B provides even more compelling evidence for this scale-dependent phenomenon. DARE merging achieves remarkable performance with virtually no degradation, -0.01\% average loss, substantially outperforming data-mixture SFT. Notably, the merged model exceeds specialist performance on several metrics, with MBPP pass@1 improving from 0.775 to 0.799. This effect suggests that task-specific weight updates can be complementary rather than competitive when sufficient model capacity exists.

\subsection{Architecture-Specific Considerations}
The results reveal architecture-dependent preferences for merging algorithms. DELLA consistently performs best for Qwen-Coder models at scale, while DARE excels for DeepSeek-Coder architectures. Internal structure and training dynamics of different model families seem to influence how well their task-specific adaptations can be merged. The consistent superiority of simpler algorithms (Linear, DARE) over more complex ones (TIES, DELLA) for smaller models indicates that merging may be counterproductive when models lack the capacity for truly separable task representations.

\subsection{Impact of Model and Dataset Scale (RQ2)}
\label{subsec:scale_analysis}

To answer \textbf{RQ2}, we performed an additional sensitive experiment in which we varied the amount of task-specific data used during SFT.  
For each model family and size, we subsampled the training splits of KodCode~\cite{xu2025kodcode} and CodeXGLUE~\cite{lu2021codexglue} to $25\%$, $50\%$, and $100\%$. 
All other hyperparameters were kept identical to those in Table~\ref{tab:hyperparameters}. Table~\ref{tab:rq2_scale} reports representative metrics for the SFT specialists.

Across all four models, HumanEval + pass@1 grows only 3-6\% when moving from the 25\% to the full 100\% subset. Prior work~\cite{yang2025scaling} has shown that synthesis oriented metrics flatten once the training corpus contains a broad coverage of idiomatic snippets, even with increased model sizes~\cite{shukor2025scalinglawsoptimaldata}. This diminishing return explains why large gains from extra data are hard to realize for code generation.

BLEU-4 almost doubles as we scale the dataset from 0.0215 to 0.0470 on DeepSeek-Coder 1.3B and 0.0351 to 0.0635 on Qwen-Coder 7B. Similar monotonic improvements with corpus size have been reported for CodeXGLUE and other comment-generation benchmarks~\cite{haldar2024analyzing}. This confirms that natural-language explanation benefits from richer lexical variety and long-tail semantic patterns that only appear in larger training sets.

Larger 7B models achieve substantially greater BLEU improvements (+0.020) compared to 2B variants despite similar pass@1 gains, confirming that increased scale primarily benefits modeling tasks through improved sample efficiency while code synthesis accuracy remains constrained by architectural design and test coverage quality~\cite{vitale2025optimizingdatasetscodesummarization}.
Model scaling effects vary by task: code generation is insensitive with additional data beyond a threshold, while code summarization shows continuous improvement as models or datasets grow. This explains why data-mixture fine-tuning works well for smaller models, but model merging becomes preferable at larger scales where specialized summarization models trained on extensive corpora transfer their improvements without degrading the already saturated generation capabilities.

\subsection{Practical Implications}
These findings have significant practical implications for deploying multi-task models under resource constraints. For organizations working with smaller models due to deployment limitations, investing in careful data mixture strategies during training will yield superior results compared to post-hoc merging approaches. Conversely, when working with larger models, the ability to merge task-specific specialists offers compelling advantages: it enables modular development where teams can independently optimize models for specific tasks before combining them, reduces training costs by avoiding redundant multi-task training, and provides flexibility to add new capabilities without retraining from scratch.
The minimal performance degradation achieved by the best approaches ($<2.1\%$ on average) validates both strategies as viable paths to multi-talented code models, with the optimal choice determined primarily by the available model scale rather than task characteristics or performance requirements.

\section{Conclusions}

In this paper, we have investigated best practices for obtaining ``small'' yet high-performing, multi-task code LLMs. Through detailed comparative assessment of data mixing and model merging techniques for two families of models across multiple parameters' scales, we have observed that combining training datasets is preferred for smaller models, while model merging is better suited for larger ones. We introduced a thorough weights differential analysis as an alternative perspective to analyze SFT, data mixing, and model merging strategies, leading to insights that can correlate with but are also complementary to performance-based examination. We have observed that, even with the same training data and hyperparameters, the patterns of model weights changes significantly across model families and scales.

Several directions remain open. First, we will extend our study beyond two tasks to a broader multi-task suite and multiple programming languages, and further sweep data mixture ratios to better characterize where the strategy can crossover. Second, we plan to go beyond evaluating existing merging methods by exploring new merging strategies that adapt to how different downstream tasks modify the model.  Finally, we will investigate theoretical links between correlation and merging success.

\begin{acks}
    This work was supported by IBM through the IBM-Rensselaer Future of Computing Research Collaboration.
    The Authors acknowledge the National Artificial Intelligence Research Resource (NAIRR) Pilot and Red Hat Research, the Mass Open Cloud (MOC), and IBM Research for contributing to this research result.
\end{acks}

\balance
\printbibliography


\appendix

\section{Model Merging Ablation Studies}

In this Section we report the ablation studies of the following hyperparameters for the four model merging strategies that we investigated in our experiments:
\begin{itemize}
    \item Linear: parameter \textbf{$weight$} in the range $$[0.1, 0.2, 0.3, 0.4, 0.5, 0.6, 0.7, 0.8, 0.9]$$
    \item TIES:
    \item DARE:
    \item DELLA:
\end{itemize}

\section*{Additional Boxplot Figures}

\begin{figure*}[htb!]
    \centering
    \begin{tabular}{cc}
    \includegraphics[width=\columnwidth]{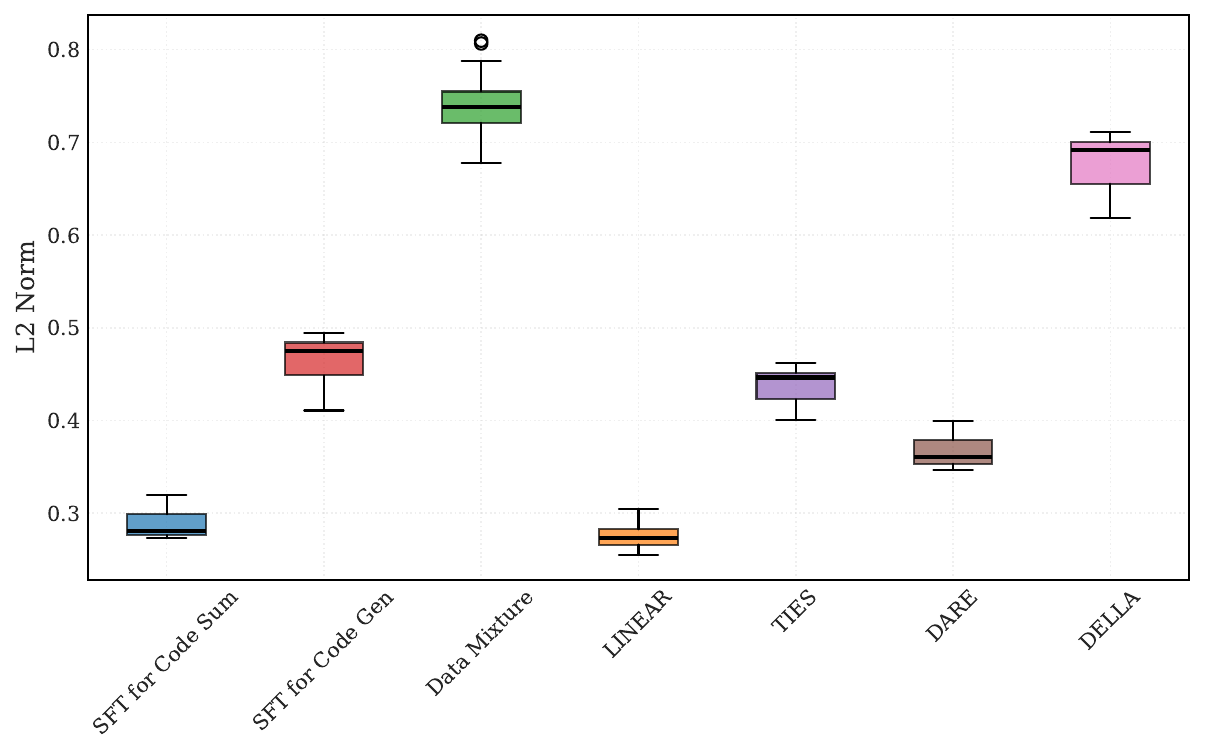} &
     \includegraphics[width=\columnwidth]{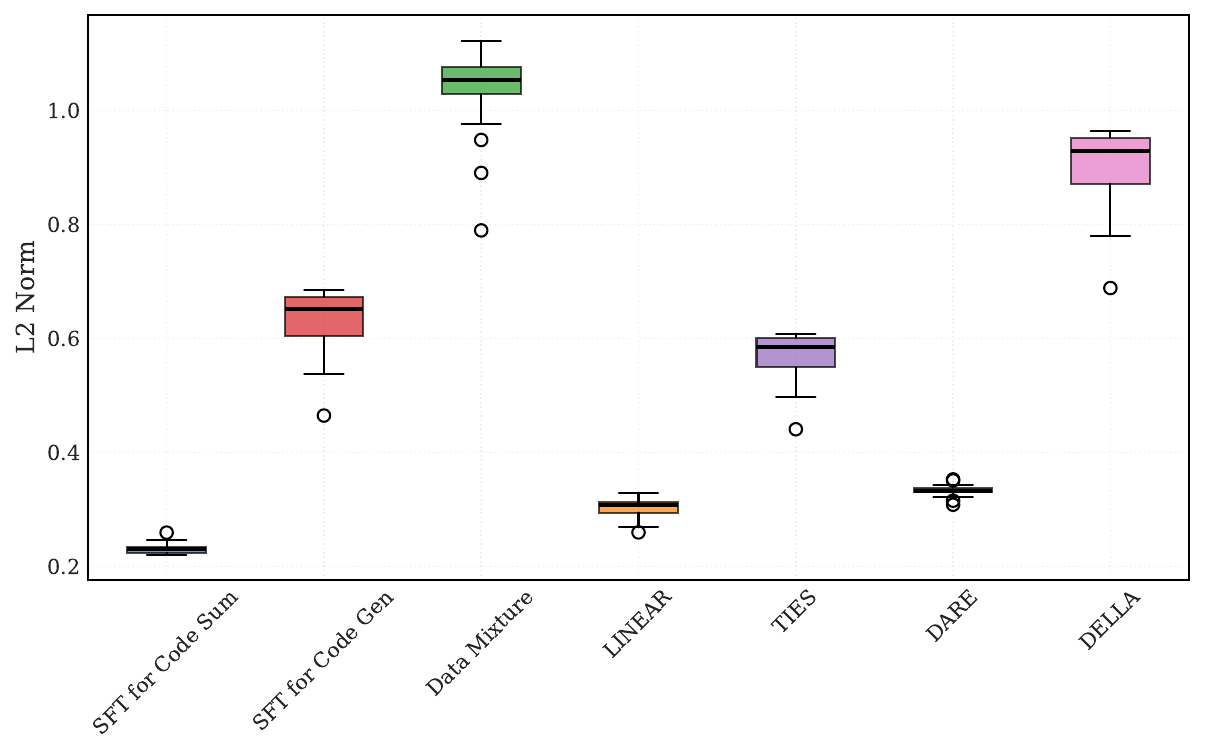}\\
    (a)  Qwen 1.5B & (b) Qwen 7B\\
    \includegraphics[width=\columnwidth]{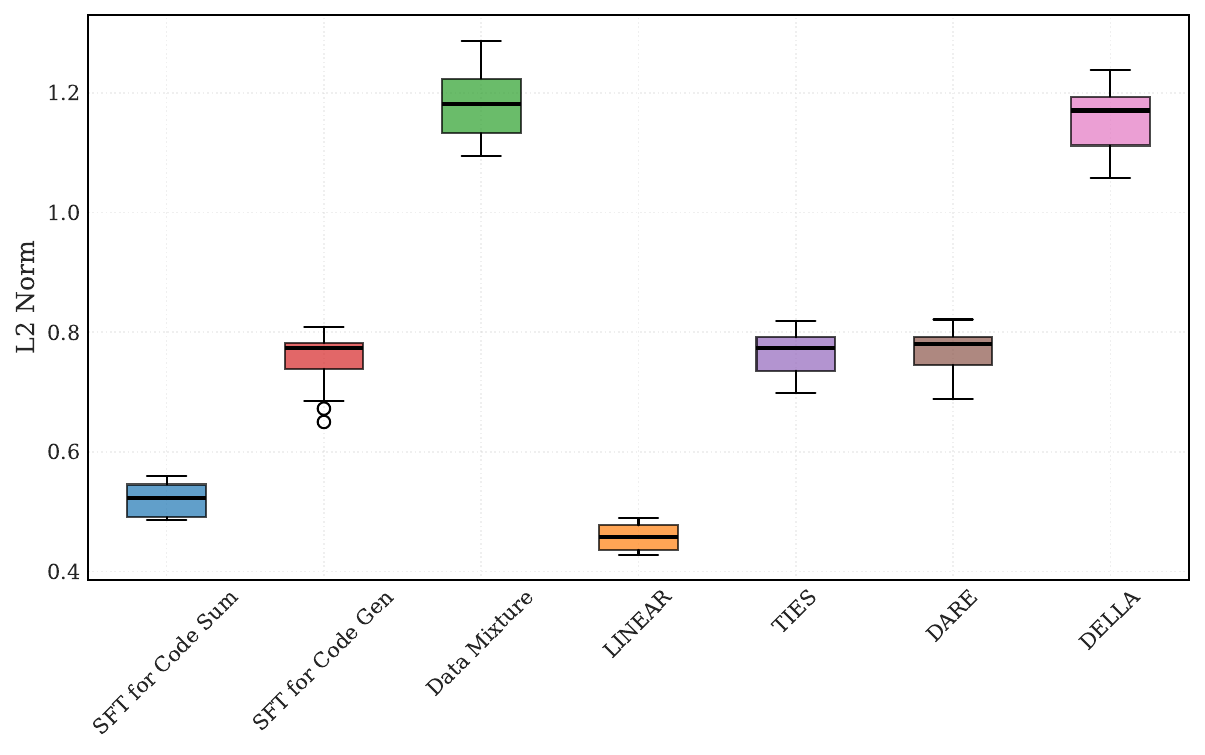}  & 
    \includegraphics[width=\columnwidth]{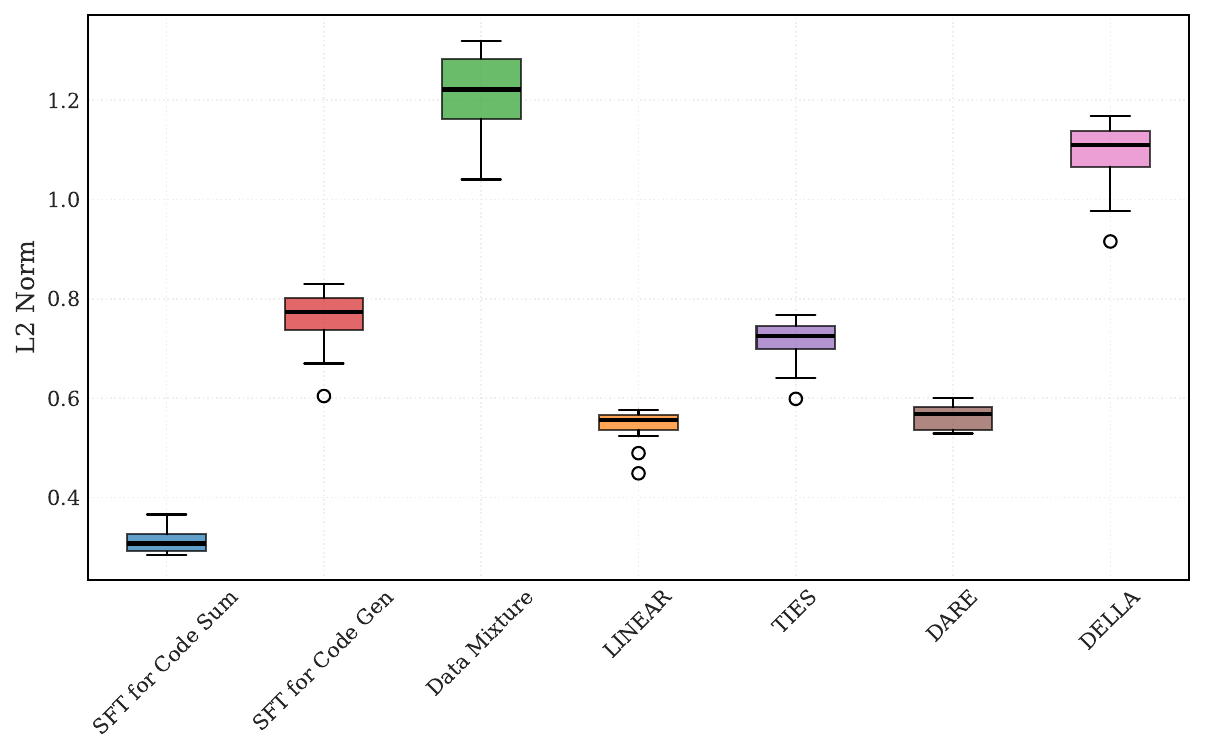}\\   
    (c) DeepSeek 1.3B & (d) DeepSeek 7B
     \end{tabular}
    \caption{Layer-wise L2 norm weight differences from base models across fine-tuning and merging approaches. Each boxplot shows parameter deviations aggregated across transformer layers for Qwen and DeepSeek models. Three supervised fine-tuning (SFT) variants are compared: CodeGen (code generation), CodeSum (code summarization), and Data-Mixture, alongside four merging strategies: Linear, TIES, DARE, and DELLA. Data-mixture SFT produces the largest weight shifts, while merged models exhibit substantially smaller deviations. Code summarization SFT shows minimal parameter changes, particularly in 7B models, suggesting this task requires fewer weight modifications from the base model.}
    \label{fig:l2_norm_boxplot_all}
\end{figure*}

The parameter deviation patterns in Fig~\ref{fig:l2_norm_boxplot_all} reveal fundamental differences in how various training strategies affect model weights across architectures and scales. Data-mixture SFT consistently exhibits the widest spread of deviations, with substantial outliers and high variance, suggesting that simultaneous multi-task learning creates conflicting optimization pressures that result in more dramatic and heterogeneous parameter changes across layers. The merging approaches demonstrate remarkably tight distributions with minimal outliers, indicating that these post-hoc strategies preserve the original parameter structure while achieving multi-task functionality through careful interpolation rather than extensive retraining. Notably, the variance patterns differ significantly between model scales: larger 7B models show more pronounced differences between approaches, with cleaner separation between SFT and merging distributions, while smaller models exhibit more overlap. The CodeSum SFT distributions are particularly revealing—their extremely low variance and absence of outliers suggest that code summarization leverages existing representational capabilities with minimal architectural disruption, contrasting sharply with CodeGen SFT which shows moderate but consistent deviations. 

\section{Ablation Study of Merge Weights}

\begin{table*}[t!] 

\centering

\resizebox{\textwidth}{!}{%
\begin{tabular}{l|c|cccc|cccc|c}
\hline
Task & \multirow{3}{*}{\centering Weight} & \multicolumn{4}{c|}{\textbf{Code Generation}} &  \multicolumn{4}{c|}{\textbf{Code Summarization}} & \\
\cline{3-10} 
Dataset & & HumanEval & HumanEval+ & MBPP & MBPP+ & \multicolumn{4}{c|}{CodeXGLUE Code-to-text} & \\
Metric & & pass@1 $\uparrow$& pass@1 $\uparrow$  & pass@1 $\uparrow$  & pass@1 $\uparrow$  & BLEU-4$\uparrow$ & chrF++$\uparrow$& ROUGE-L$\uparrow$ & METEOR $\uparrow$& AVG \% wrt SFT  \\
\hline
\multicolumn{2}{c}{} & \multicolumn{8}{c}{\textit{Qwen2.5-Coder 1.5B}} &  \\
\hline
SFT-CodeGen/Sum & - & 0.756(0\%) & 0.707(0\%) & 0.759(0\%) & 0.656(0\%) & 0.0495(0\%) & 0.3213(0\%) & 0.3279(0\%) & 0.3091(0\%) & 0.00\% \\
\hline
 \multirow{9}{*}{\centering Linear}& 0.1/0.9 & 0.707(-6.48\%) & 0.665(-5.94\%) & 0.757(-0.26\%) & 0.659(+0.46\%) & 0.0303(-38.79\%) & 0.252(-21.57\%) & 0.2637(-19.58\%) & 0.2294(-25.78\%) & -14.74\% \\
 & 0.2/0.8& 0.622(-17.72\%) & 0.573(-18.95\%) & 0.722(-4.87\%) & 0.632(-3.66\%) & 0.0451(-8.89\%) & 0.317(-1.34\%) & 0.3203(-2.32\%) & 0.3046(-1.46\%) & -7.40\% \\
 & 0.3/0.7 & \textbf{0.671(-11.24\%)} & \textbf{0.598(-15.42\%)} & \textbf{0.749(-1.32\%)} & \textbf{0.651(-0.76\%)} & \textbf{0.0434(-12.32\%)} & \textbf{0.306(-4.76\%)} & \textbf{0.3086(-5.89\%)} & \textbf{0.2912(-5.79\%)} & \textbf{-7.19\%} \\
 & 0.4/0.6 & 0.671(-11.24\%) & 0.616(-12.87\%) & 0.762(+0.40\%) & 0.664(+1.22\%) & 0.0383(-22.63\%) & 0.2836(-11.73\%) & 0.2883(-12.08\%) & 0.2662(-13.88\%) & -10.35\% \\
 & 0.5/0.5 & 0.707(-6.48\%) & 0.665(-5.94\%) & 0.757(-0.26\%) & 0.659(+0.46\%) & 0.0303(-38.79\%) & 0.252(-21.57\%) & 0.2636(-19.61\%) & 0.2294(-25.78\%) & -14.75\% \\
 & 0.6/0.4 & 0.738(-2.38\%) & 0.652(-7.78\%) & 0.77(+1.45\%) & 0.669(+1.98\%) & 0.0223(-54.95\%) & 0.2336(-27.30\%) & 0.2301(-29.83\%) & 0.1993(-35.52\%) & -19.29\% \\
 & 0.7/0.3& 0.72(-4.76\%) & 0.671(-5.09\%) & 0.775(+2.11\%) & 0.667(+1.68\%) & 0.012(-75.76\%) & 0.2068(-35.64\%) & 0.1863(-43.18\%) & 0.1488(-51.86\%) & -26.56\% \\
 & 0.8/0.2 & 0.713(-5.69\%) & 0.677(-4.24\%) & 0.77(+1.45\%) & 0.667(+1.68\%) & 0.0056(-88.69\%) & 0.1998(-37.82\%) & 0.1542(-52.97\%) & 0.1223(-60.43\%) & -30.84\% \\
 & 0.9/0.1 & 0.726(-3.97\%) & 0.689(-2.55\%) & 0.772(+1.71\%) & 0.664(+1.22\%) & 0.0025(-94.95\%) & 0.1935(-39.78\%) & 0.1365(-58.37\%) & 0.1091(-64.70\%) & -32.67\% \\
\hline
\multirow{9}{*}{\centering TIES} & 0.1/0.9 & 0.646(-14.55\%) & 0.585(-17.26\%) & 0.757(-0.26\%) & 0.653(-0.46\%) & 0.0113(-77.17\%) & 0.2211(-31.19\%) & 0.1742(-46.87\%) & 0.1579(-48.92\%) & -29.58\% \\
 & 0.2/0.8 & \textbf{0.659(-12.83\%)} & \textbf{0.591(-16.41\%)} & \textbf{0.738(-2.77\%)} & \textbf{0.63(-3.96\%)} & \textbf{0.0432(-12.73\%)} & \textbf{0.3177(-1.12\%)} & \textbf{0.3114(-5.03\%)} & \textbf{0.3059(-1.04\%)} & \textbf{-6.99\%} \\
 & 0.3/0.7 & 0.652(-13.76\%) & 0.591(-16.41\%) & 0.754(-0.66\%) & 0.635(-3.20\%) & 0.0372(-24.85\%) & 0.3024(-5.88\%) & 0.2894(-11.74\%) & 0.2883(-6.73\%) & -10.40\% \\
 & 0.4/0.6 & 0.646(-14.55\%) & 0.598(-15.42\%) & 0.751(-1.05\%) & 0.635(-3.20\%) & 0.0199(-59.80\%) & 0.2461(-23.40\%) & 0.212(-35.35\%) & 0.2049(-33.71\%) & -23.31\% \\
 & 0.5/0.5 & 0.646(-14.55\%) & 0.585(-17.26\%) & 0.757(-0.26\%) & 0.653(-0.46\%) & 0.0113(-77.17\%) & 0.2211(-31.19\%) & 0.1741(-46.90\%) & 0.1579(-48.92\%) & -29.59\% \\
 & 0.6/0.4 & 0.671(-11.24\%) & 0.622(-12.02\%) & 0.762(+0.40\%) & 0.667(+1.68\%) & 0.0098(-80.20\%) & 0.2137(-33.49\%) & 0.1643(-49.89\%) & 0.1475(-52.28\%) & -29.63\% \\
 & 0.7/0.3 & 0.677(-10.45\%) & 0.634(-10.33\%) & 0.767(+1.05\%) & 0.667(+1.68\%) & 0.0082(-83.43\%) & 0.209(-34.95\%) & 0.1579(-51.85\%) & 0.1415(-54.22\%) & -30.31\% \\
 & 0.8/0.2 & 0.683(-9.66\%) & 0.64(-9.48\%) & 0.759(0.00\%) & 0.664(+1.22\%) & 0.008(-83.84\%) & 0.2089(-34.98\%) & 0.1564(-52.30\%) & 0.1412(-54.32\%) & -30.42\% \\
 & 0.9/0.1 & 0.683(-9.66\%) & 0.64(-9.48\%) & 0.762(+0.40\%) & 0.667(+1.68\%) & 0.0079(-84.04\%) & 0.2088(-35.01\%) & 0.1552(-52.67\%) & 0.1415(-54.22\%) & -30.38\% \\
\hline
\multirow{9}{*}{\centering DARE} & 0.1/0.9 & 0.732(-3.17\%) & 0.665(-5.94\%) & 0.735(-3.16\%) & 0.653(-0.46\%) & 0.0161(-67.47\%) & 0.2718(-15.41\%) & 0.2128(-35.10\%) & 0.2421(-21.68\%) & -19.05\% \\
 & 0.2/0.8 & \textbf{0.616(-18.52\%)} & \textbf{0.567(-19.80\%)} & \textbf{0.725(-4.48\%)} & \textbf{0.638(-2.74\%)} & \textbf{0.0468(-5.45\%)} & \textbf{0.3157(-1.74\%)} & \textbf{0.3205(-2.26\%)} & \textbf{0.3033(-1.88\%)} & \textbf{-7.11\%} \\
 & 0.3/0.7 & 0.665(-12.04\%) & 0.585(-17.26\%) & 0.754(-0.66\%) & 0.659(+0.46\%) & 0.0427(-13.74\%) & 0.3035(-5.54\%) & 0.3067(-6.47\%) & 0.2885(-6.66\%) & -7.74\% \\
 & 0.4/0.6 & 0.689(-8.86\%) & 0.634(-10.33\%) & 0.759(0.00\%) & 0.664(+1.22\%) & 0.0365(-26.26\%) & 0.2757(-14.19\%) & 0.2843(-13.30\%) & 0.2576(-16.66\%) & -11.05\% \\
 & 0.5/0.5 & 0.707(-6.48\%) & 0.659(-6.79\%) & 0.754(-0.66\%) & 0.664(+1.22\%) & 0.0306(-38.18\%) & 0.2515(-21.72\%) & 0.263(-19.79\%) & 0.2291(-25.88\%) & -14.79\% \\
 & 0.6/0.4 & 0.713(-5.69\%) & 0.646(-8.63\%) & 0.77(+1.45\%) & 0.669(+1.98\%) & 0.0226(-54.34\%) & 0.2331(-27.45\%) & 0.2336(-28.76\%) & 0.2004(-35.17\%) & -19.58\% \\
 & 0.7/0.3 & 0.707(-6.48\%) & 0.652(-7.78\%) & 0.77(+1.45\%) & 0.669(+1.98\%) & 0.0135(-72.73\%) & 0.2099(-34.67\%) & 0.1941(-40.81\%) & 0.1567(-49.30\%) & -26.04\% \\
 & 0.8/0.2 & 0.707(-6.48\%) & 0.671(-5.09\%) & 0.77(+1.45\%) & 0.661(+0.76\%) & 0.005(-89.90\%) & 0.1975(-38.53\%) & 0.1516(-53.77\%) & 0.1193(-61.40\%) & -31.62\% \\
 & 0.9/0.1 & 0.726(-3.97\%) & 0.689(-2.55\%) & 0.78(+2.77\%) & 0.667(+1.68\%) & 0.0023(-95.35\%) & 0.1927(-40.02\%) & 0.1345(-58.98\%) & 0.1078(-65.12\%) & -32.69\% \\
\hline
\multirow{9}{*}{\centering DELLA} & 0.1/0.9 & 0.591(-21.83\%) & 0.549(-22.35\%) & 0.733(-3.43\%) & 0.638(-2.74\%) & 0.0056(-88.69\%) & 0.2023(-37.04\%) & 0.1456(-55.60\%) & 0.1302(-57.88\%) & -36.19\% \\
 & 0.2/0.8 & \textbf{0.561(-25.79\%)} & \textbf{0.518(-26.73\%)} & \textbf{0.714(-5.93\%)} & \textbf{0.611(-6.86\%)} & \textbf{0.0297(-40.00\%)} & \textbf{0.2843(-11.52\%)} & \textbf{0.2684(-18.15\%)} & \textbf{0.265(-14.27\%)} & \textbf{-18.66\%} \\
 & 0.3/0.7 & 0.573(-24.21\%) & 0.524(-25.88\%) & 0.712(-6.19\%) & 0.622(-5.18\%) & 0.0183(-63.03\%) & 0.2422(-24.62\%) & 0.2041(-37.76\%) & 0.2033(-34.23\%) & -27.64\% \\
 & 0.4/0.6 & 0.573(-24.21\%) & 0.518(-26.73\%) & 0.722(-4.87\%) & 0.622(-5.18\%) & 0.0071(-85.66\%) & 0.2066(-35.70\%) & 0.1527(-53.43\%) & 0.1373(-55.58\%) & -36.42\% \\
 & 0.5/0.5 & 0.634(-16.14\%) & 0.598(-15.42\%) & 0.722(-4.87\%) & 0.622(-5.18\%) & 0.0052(-89.49\%) & 0.2002(-37.69\%) & 0.1423(-56.60\%) & 0.1284(-58.46\%) & -35.48\% \\
 & 0.6/0.4 & 0.646(-14.55\%) & 0.598(-15.42\%) & 0.746(-1.71\%) & 0.643(-1.98\%) & 0.0056(-88.69\%) & 0.2011(-37.41\%) & 0.1426(-56.51\%) & 0.1308(-57.68\%) & -34.24\% \\
 & 0.7/0.3 & 0.671(-11.24\%) & 0.628(-11.17\%) & 0.746(-1.71\%) & 0.632(-3.66\%) & 0.0044(-91.11\%) & 0.1969(-38.72\%) & 0.1379(-57.94\%) & 0.1231(-60.17\%) & -34.47\% \\
 & 0.8/0.2 & 0.646(-14.55\%) & 0.604(-14.57\%) & 0.746(-1.71\%) & 0.638(-2.74\%) & 0.005(-89.90\%) & 0.1977(-38.47\%) & 0.1382(-57.85\%) & 0.1255(-59.40\%) & -34.90\% \\
 & 0.9/0.1 & 0.665(-12.04\%) & 0.622(-12.02\%) & 0.749(-1.32\%) & 0.653(-0.46\%) & 0.0041(-91.72\%) & 0.1954(-39.18\%) & 0.1353(-58.74\%) & 0.1206(-60.98\%) & -34.56\% \\
\hline

\end{tabular}}
\caption{Qwen2.5-Coder 1.5B ablation study results showing performance across different merging strategies (Linear, TIES, DARE, DELLA) with varying weight ratios. Each row represents a different CodeGen/CodeSum weight combination, with performance metrics for code generation tasks (HumanEval, HumanEval+, MBPP, MBPP+) and code summarization tasks (CodeXGLUE). Percentage changes are relative to the SFT baseline. The rightmost column shows the average percentage change across all metrics.}
\label{tab:ablation_results_qwc15b}
\end{table*}

\begin{table*}[t!] 
\centering

\resizebox{\textwidth}{!}{%
\begin{tabular}{l|c|cccc|cccc|c}
\hline
Task & \multirow{3}{*}{\centering Weight} & \multicolumn{4}{c|}{\textbf{Code Generation}} &  \multicolumn{4}{c|}{\textbf{Code Summarization}} & \\
\cline{3-10} 
Dataset & & HumanEval & HumanEval+ & MBPP & MBPP+ & \multicolumn{4}{c|}{CodeXGLUE Code-to-text} & \\
Metric & & pass@1 $\uparrow$& pass@1 $\uparrow$  & pass@1 $\uparrow$  & pass@1 $\uparrow$  & BLEU-4$\uparrow$ & chrF++$\uparrow$& ROUGE-L$\uparrow$ & METEOR $\uparrow$& AVG \% wrt SFT  \\
\hline
\multicolumn{2}{c}{} & \multicolumn{8}{c}{\textit{Qwen2.5-Coder 7B}} &  \\
\hline
SFT-CodeGen/Sum & - & 0.909(0\%) & 0.866(0\%) & 0.874(0\%) & 0.741(0\%) & 0.0635(0\%) & 0.3356(0\%) & 0.3436(0\%) & 0.3243(0\%) & 0.00\% \\
\hline
\multirow{9}{*}{\centering Linear} & 0.1/0.9 & 0.866(-4.73\%) & 0.817(-5.66\%) & 0.854(-2.73\%) & 0.735(-1.08\%) & 0.0496(-21.89\%) & 0.2936(-12.51\%) & 0.3172(-7.95\%) & 0.2735(-16.08\%) & -9.08\% \\
 & \textbf{0.2/0.8} & \textbf{0.811(-10.78\%)} & \textbf{0.762(-12.01\%)} & \textbf{0.815(-7.18\%)} & \textbf{0.704(-5.25\%)} & \textbf{0.0613(-3.46\%)} & \textbf{0.3339(-0.51\%)} & \textbf{0.3452(+0.17\%)} & \textbf{0.3231(-0.86\%)} & \textbf{-4.98\%} \\
 & 0.3/0.7 & 0.835(-8.14\%) & 0.787(-9.12\%) & 0.841(-4.21\%) & 0.712(-4.17\%) & 0.0601(-5.35\%) & 0.3298(-1.73\%) & 0.3419(-0.78\%) & 0.3175(-2.58\%) & -4.51\% \\
 & 0.4/0.6 & 0.866(-4.73\%) & 0.811(-6.35\%) & 0.839(-4.44\%) & 0.709(-4.58\%) & 0.0558(-12.13\%) & 0.317(-5.54\%) & 0.3326(-3.48\%) & 0.3015(-7.49\%) & -6.09\% \\
 & 0.5/0.5 & 0.866(-4.73\%) & 0.817(-5.66\%) & 0.854(-2.73\%) & 0.735(-1.08\%) & 0.0496(-21.89\%) & 0.2936(-12.51\%) & 0.3171(-7.98\%) & 0.2735(-16.08\%) & -9.08\% \\
 & 0.6/0.4 & 0.878(-3.41\%) & 0.817(-5.66\%) & 0.849(-3.30\%) & 0.72(-3.10\%) & 0.0414(-34.80\%) & 0.2759(-17.79\%) & 0.3025(-12.22\%) & 0.2525(-22.52\%) & -12.85\% \\
 & 0.7/0.3 & 0.896(-1.43\%) & 0.848(-2.08\%) & 0.857(-2.39\%) & 0.735(-1.08\%) & 0.0343(-45.98\%) & 0.2784(-17.04\%) & 0.291(-15.55\%) & 0.253(-22.37\%) & -13.49\% \\
 & 0.8/0.2 & 0.909(0.00\%) & 0.86(-0.69\%) & 0.873(-0.57\%) & 0.738(-0.67\%) & 0.0321(-49.45\%) & 0.3139(-6.47\%) & 0.287(-16.72\%) & 0.2978(-8.62\%) & -10.40\% \\
 & 0.9/0.1 & 0.909(0.00\%) & 0.848(-2.08\%) & 0.868(-1.14\%) & 0.733(-1.35\%) & 0.0316(-50.24\%) & 0.3351(-0.15\%) & 0.2822(-18.11\%) & 0.329(+0.95\%) & -9.01\% \\
\hline
\multirow{9}{*}{\centering TIES} & 0.1/0.9 & 0.921(+1.32\%) & 0.866(0.00\%) & 0.87(-0.91\%) & 0.728(-2.02\%) & 0.0501(-21.10\%) & 0.3067(-8.61\%) & 0.3192(-7.37\%) & 0.2879(-11.66\%) & -6.29\% \\
 & 0.2/0.8 & 0.823(-9.46\%) & 0.787(-9.12\%) & 0.847(-3.53\%) & 0.72(-3.10\%) & 0.0609(-4.09\%) & 0.336(+0.12\%) & 0.3465(+0.55\%) & 0.3258(-0.03\%) & -3.58\% \\
 & 0.3/0.7 & 0.866(-4.73\%) & 0.811(-6.35\%) & 0.862(-1.82\%) & 0.728(-2.02\%) & 0.0593(-6.61\%) & 0.3333(-0.69\%) & 0.3428(-0.52\%) & 0.3215(-1.35\%) & -3.01\% \\
 & 0.4/0.6 & 0.921(+1.32\%) & 0.86(-0.69\%) & 0.873(-0.57\%) & 0.735(-1.08\%) & 0.0533(-16.06\%) & 0.3127(-6.82\%) & 0.3269(-5.14\%) & 0.2953(-9.39\%) & -4.80\% \\
 & 0.5/0.5 & 0.921(+1.32\%) & 0.866(0.00\%) & 0.87(-0.91\%) & 0.728(-2.02\%) & 0.0501(-21.10\%) & 0.3067(-8.61\%) & 0.3193(-7.34\%) & 0.2879(-11.66\%) & -6.29\% \\
 & 0.6/0.4 & 0.921(+1.32\%) & 0.866(0.00\%) & 0.862(-1.82\%) & 0.725(-2.42\%) & 0.0487(-23.31\%) & 0.3088(-7.99\%) & 0.3175(-7.86\%) & 0.2902(-10.95\%) & -6.63\% \\
 & 0.7/0.3 & 0.921(+1.32\%) & 0.866(0.00\%) & 0.862(-1.82\%) & 0.728(-2.02\%) & 0.0498(-21.57\%) & 0.312(-7.03\%) & 0.3201(-7.11\%) & 0.294(-9.79\%) & -6.00\% \\
 & 0.8/0.2 & 0.927(+1.98\%) & 0.866(0.00\%) & 0.86(-2.05\%) & 0.733(-1.35\%) & 0.0518(-18.43\%) & 0.3181(-5.21\%) & 0.3251(-5.66\%) & 0.3017(-7.43\%) & -4.77\% \\
 & \textbf{0.9/0.1} & \textbf{0.921(+1.32\%)} & \textbf{0.86(-0.69\%)} & \textbf{0.86(-2.05\%)} & \textbf{0.73(-1.75\%)} & \textbf{0.0536(-15.59\%)} & \textbf{0.3234(-3.64\%)} & \textbf{0.3285(-4.67\%)} & \textbf{0.3084(-5.37\%)} & \textbf{-4.06\%} \\
\hline
\multirow{9}{*}{\centering DARE} & 0.1/0.9 & 0.915(+0.66\%) & 0.848(-2.08\%) & 0.881(+0.34\%) & 0.746(+0.40\%) & 0.0317(-50.08\%) & 0.3089(-7.96\%) & 0.2856(-17.12\%) & 0.2913(-10.62\%) & -10.81\% \\
 & \textbf{0.2/0.8} & \textbf{0.829(-8.80\%)} & \textbf{0.78(-9.93\%)} & \textbf{0.815(-7.18\%)} & \textbf{0.704(-5.25\%)} & \textbf{0.0614(-3.31\%)} & \textbf{0.334(-0.48\%)} & \textbf{0.3459(+0.38\%)} & \textbf{0.3235(-0.74\%)} & \textbf{-4.41\%} \\
 & 0.3/0.7 & 0.841(-7.48\%) & 0.799(-7.74\%) & 0.844(-3.87\%) & 0.722(-2.83\%) & 0.0579(-8.82\%) & 0.3243(-3.37\%) & 0.338(-1.92\%) & 0.3106(-4.69\%) & -5.09\% \\
 & 0.4/0.6 & 0.884(-2.75\%) & 0.829(-4.27\%) & 0.86(-2.05\%) & 0.72(-3.10\%) & 0.0538(-15.28\%) & 0.309(-7.93\%) & 0.3279(-4.85\%) & 0.2917(-10.49\%) & -6.34\% \\
 & 0.5/0.5 & 0.878(-3.41\%) & 0.829(-4.27\%) & 0.849(-3.30\%) & 0.725(-2.42\%) & 0.0496(-21.89\%) & 0.2942(-12.34\%) & 0.3175(-7.86\%) & 0.2741(-15.89\%) & -8.92\% \\
 & 0.6/0.4 & 0.902(-0.77\%) & 0.835(-3.58\%) & 0.854(-2.73\%) & 0.725(-2.42\%) & 0.0444(-30.08\%) & 0.2808(-16.33\%) & 0.3076(-10.74\%) & 0.2588(-20.59\%) & -10.91\% \\
 & 0.7/0.3 & 0.915(+0.66\%) & 0.854(-1.39\%) & 0.849(-3.30\%) & 0.73(-1.75\%) & 0.0383(-39.69\%) & 0.2779(-17.19\%) & 0.2979(-13.55\%) & 0.2538(-22.12\%) & -12.29\% \\
 & 0.8/0.2 & 0.915(+0.66\%) & 0.86(-0.69\%) & 0.886(+0.91\%) & 0.751(+1.08\%) & 0.0307(-51.65\%) & 0.3259(-2.89\%) & 0.2833(-17.79\%) & 0.3164(-2.92\%) & -9.16\% \\
 & 0.9/0.1 & 0.909(0.00\%) & 0.854(-1.39\%) & 0.873(-0.57\%) & 0.741(-0.27\%) & 0.0302(-52.44\%) & 0.3327(-0.86\%) & 0.2793(-18.95\%) & 0.3259(0.00\%) & -9.31\% \\
\hline
\multirow{9}{*}{\centering DELLA} & 0.1/0.9 & 0.896(-1.43\%) & 0.841(-2.89\%) & 0.857(-2.39\%) & 0.738(-0.67\%) & 0.0578(-8.98\%) & 0.3388(+0.95\%) & 0.3392(-1.57\%) & 0.327(+0.34\%) & -2.08\% \\
 & 0.2/0.8 & 0.866(-4.73\%) & 0.817(-5.66\%) & 0.847(-3.53\%) & 0.728(-2.02\%) & 0.059(-7.09\%) & 0.3338(-0.54\%) & 0.3456(+0.29\%) & 0.3231(-0.86\%) & -3.02\% \\
 & 0.3/0.7 & 0.86(-5.39\%) & 0.811(-6.35\%) & 0.852(-2.96\%) & 0.733(-1.35\%) & 0.0592(-6.77\%) & 0.3365(+0.27\%) & 0.3449(+0.09\%) & 0.326(+0.03\%) & -2.80\% \\
 & 0.4/0.6 & 0.878(-3.41\%) & 0.823(-4.97\%) & 0.849(-3.30\%) & 0.733(-1.35\%) & 0.059(-7.09\%) & 0.3396(+1.19\%) & 0.3444(-0.06\%) & 0.3292(+1.01\%) & -2.25\% \\
 & \textbf{0.5/0.5} & \textbf{0.902(-0.77\%)} & \textbf{0.841(-2.89\%)} & \textbf{0.849(-3.30\%)} & \textbf{0.735(-1.08\%)} & \textbf{0.0578(-8.98\%)} & \textbf{0.3398(+1.25\%)} & \textbf{0.3391(-1.60\%)} & \textbf{0.3284(+0.77\%)} & \textbf{-2.07\%} \\
 & 0.6/0.4 & 0.909(0.00\%) & 0.841(-2.89\%) & 0.849(-3.30\%) & 0.735(-1.08\%) & 0.0574(-9.61\%) & 0.3379(+0.69\%) & 0.3385(-1.77\%) & 0.3259(0.00\%) & -2.24\% \\
 & 0.7/0.3 & 0.902(-0.77\%) & 0.841(-2.89\%) & 0.852(-2.96\%) & 0.738(-0.67\%) & 0.0573(-9.76\%) & 0.3383(+0.80\%) & 0.3367(-2.29\%) & 0.3263(+0.12\%) & -2.30\% \\
 & 0.8/0.2 & 0.902(-0.77\%) & 0.841(-2.89\%) & 0.852(-2.96\%) & 0.728(-2.02\%) & 0.0573(-9.76\%) & 0.3391(+1.04\%) & 0.3369(-2.23\%) & 0.3274(+0.46\%) & -2.39\% \\
 & 0.9/0.1 & 0.902(-0.77\%) & 0.841(-2.89\%) & 0.849(-3.30\%) & 0.73(-1.75\%) & 0.0576(-9.29\%) & 0.3404(+1.43\%) & 0.3374(-2.09\%) & 0.3288(+0.89\%) & -2.22\% \\
\hline

\end{tabular}}
\caption{Qwen2.5-Coder 7B ablation study results showing performance across different merging strategies (Linear, TIES, DARE, DELLA) with varying weight ratios. Each row represents a different CodeGen/CodeSum weight combination, with performance metrics for code generation tasks (HumanEval, HumanEval+, MBPP, MBPP+) and code summarization tasks (CodeXGLUE). Percentage changes are relative to the SFT baseline. The rightmost column shows the average percentage change across all metrics.}
\label{tab:ablation_results_qwc7b}
\end{table*}

\begin{table*}[t!] 
\centering

\resizebox{\textwidth}{!}{%
\begin{tabular}{l|c|cccc|cccc|c}
\hline
Task & \multirow{3}{*}{\centering Weight} & \multicolumn{4}{c|}{\textbf{Code Generation}} &  \multicolumn{4}{c|}{\textbf{Code Summarization}} & \\
\cline{3-10} 
Dataset & & HumanEval & HumanEval+ & MBPP & MBPP+ & \multicolumn{4}{c|}{CodeXGLUE Code-to-text} & \\
Metric & & pass@1 $\uparrow$& pass@1 $\uparrow$  & pass@1 $\uparrow$  & pass@1 $\uparrow$  & BLEU-4$\uparrow$ & chrF++$\uparrow$& ROUGE-L$\uparrow$ & METEOR $\uparrow$& AVG \% wrt SFT  \\
\hline
\multicolumn{2}{c}{} & \multicolumn{8}{c}{\textit{DeepSeek Coder 1.3B}} &  \\
\hline
SFT-CodeGen/Sum & - & 0.707(0\%) & 0.671(0\%) & 0.629(0\%) & 0.548(0\%) & 0.047(0\%) & 0.3214(0\%) & 0.324(0\%) & 0.3083(0\%) & 0.00\% \\
\hline
\multirow{9}{*}{\centering Linear} & 0.1/0.9 & 0.677(-4.24\%) & 0.61(-9.09\%) & 0.667(+5.87\%) & 0.571(+4.20\%) & 0.0397(-15.53\%) & 0.3051(-5.01\%) & 0.2716(-16.25\%) & 0.2998(-2.91\%) & -5.37\% \\
 & 0.2/0.8 & 0.591(-16.41\%) & 0.537(-19.97\%) & 0.616(-2.22\%) & 0.54(-1.46\%) & 0.0473(+0.64\%) & 0.3304(+2.86\%) & 0.3188(-1.70\%) & 0.3197(+3.53\%) & -4.34\% \\
 & \textbf{0.3/0.7} & \textbf{0.628(-11.17\%)} & \textbf{0.561(-16.39\%)} & \textbf{0.646(+2.54\%)} & \textbf{0.558(+1.82\%)} & \textbf{0.046(-2.13\%)} & \textbf{0.3309(+3.02\%)} & \textbf{0.3127(-3.58\%)} & \textbf{0.3208(+3.89\%)} & \textbf{-2.75\%} \\
 & 0.4/0.6 & 0.628(-11.17\%) & 0.573(-14.61\%) & 0.672(+6.67\%) & 0.582(+6.20\%) & 0.0447(-4.89\%) & 0.3223(+0.34\%) & 0.2963(-8.63\%) & 0.3141(+1.72\%) & -3.05\% \\
 & 0.5/0.5 & 0.677(-4.24\%) & 0.61(-9.09\%) & 0.667(+5.87\%) & 0.571(+4.20\%) & 0.0397(-15.53\%) & 0.3051(-5.01\%) & 0.2716(-16.25\%) & 0.2998(-2.91\%) & -5.37\% \\
 & 0.6/0.4 & 0.671(-5.09\%) & 0.616(-8.20\%) & 0.653(+3.65\%) & 0.563(+2.74\%) & 0.031(-34.04\%) & 0.2624(-18.31\%) & 0.2189(-32.50\%) & 0.2627(-14.93\%) & -13.33\% \\
 & 0.7/0.3 & 0.671(-5.09\%) & 0.622(-7.30\%) & 0.651(+3.33\%) & 0.571(+4.20\%) & 0.029(-38.30\%) & 0.2902(-9.65\%) & 0.2403(-25.90\%) & 0.2849(-7.74\%) & -10.81\% \\
 & 0.8/0.2 & 0.665(-5.94\%) & 0.622(-7.30\%) & 0.648(+2.86\%) & 0.569(+3.83\%) & 0.0133(-71.70\%) & 0.2506(-21.98\%) & 0.19(-41.41\%) & 0.2096(-32.12\%) & -21.72\% \\
 & 0.9/0.1 & 0.695(-1.70\%) & 0.652(-2.83\%) & 0.648(+2.86\%) & 0.556(+1.46\%) & 0.0075(-84.04\%) & 0.2265(-29.48\%) & 0.1537(-52.61\%) & 0.1834(-40.61\%) & -25.87\% \\
\hline
\multirow{9}{*}{\centering TIES} & 0.1/0.9 & 0.628(-11.17\%) & 0.591(-11.92\%) & 0.661(+4.92\%) & 0.574(+4.74\%) & 0.0372(-20.85\%) & 0.2877(-10.43\%) & 0.2524(-22.17\%) & 0.2844(-7.90\%) & -9.35\% \\
 & 0.2/0.8 & 0.591(-16.41\%) & 0.549(-18.18\%) & 0.624(-0.95\%) & 0.542(-1.09\%) & 0.045(-4.26\%) & 0.3241(+0.90\%) & 0.3068(-5.40\%) & 0.3147(+1.91\%) & -5.43\% \\
 & \textbf{0.3/0.7} & \textbf{0.61(-13.72\%)} & \textbf{0.561(-16.39\%)} & \textbf{0.622(-1.27\%)} & \textbf{0.545(-0.55\%)} & \textbf{0.0452(-3.83\%)} & \textbf{0.3213(+0.03\%)} & \textbf{0.3007(-7.28\%)} & \textbf{0.3129(+1.33\%)} & \textbf{-5.21\%} \\
 & 0.4/0.6 & 0.64(-9.48\%) & 0.604(-9.99\%) & 0.659(+4.60\%) & 0.566(+3.28\%) & 0.0413(-12.13\%) & 0.3096(-3.61\%) & 0.2821(-13.01\%) & 0.3032(-1.81\%) & -5.27\% \\
 & 0.5/0.5 & 0.628(-11.17\%) & 0.591(-11.92\%) & 0.661(+4.92\%) & 0.574(+4.74\%) & 0.0372(-20.85\%) & 0.2877(-10.43\%) & 0.2525(-22.14\%) & 0.2844(-7.90\%) & -9.34\% \\
 & 0.6/0.4 & 0.652(-7.78\%) & 0.604(-9.99\%) & 0.661(+4.92\%) & 0.566(+3.28\%) & 0.0318(-32.34\%) & 0.2638(-17.87\%) & 0.2219(-31.58\%) & 0.2628(-14.90\%) & -13.28\% \\
 & 0.7/0.3 & 0.659(-6.79\%) & 0.61(-9.09\%) & 0.667(+5.87\%) & 0.571(+4.20\%) & 0.0295(-37.23\%) & 0.2595(-19.21\%) & 0.2147(-33.80\%) & 0.2582(-16.39\%) & -14.05\% \\
 & 0.8/0.2 & 0.671(-5.09\%) & 0.628(-6.41\%) & 0.651(+3.33\%) & 0.569(+3.83\%) & 0.0289(-38.51\%) & 0.2589(-19.40\%) & 0.2135(-34.17\%) & 0.2573(-16.68\%) & -14.14\% \\
 & 0.9/0.1 & 0.659(-6.79\%) & 0.616(-8.20\%) & 0.667(+5.87\%) & 0.579(+5.66\%) & 0.0291(-38.09\%) & 0.2576(-19.80\%) & 0.2124(-34.51\%) & 0.2565(-16.94\%) & -14.10\% \\
\hline
\multirow{9}{*}{\centering DARE} & 0.1/0.9 & 0.665(-5.94\%) & 0.604(-9.99\%) & 0.651(+3.33\%) & 0.553(+0.91\%) & 0.0089(-81.06\%) & 0.2337(-27.24\%) & 0.1591(-50.94\%) & 0.2096(-32.12\%) & -25.38\% \\
 & 0.2/0.8 & 0.585(-17.26\%) & 0.53(-21.01\%) & 0.632(+0.32\%) & 0.556(+1.46\%) & 0.0469(-0.21\%) & 0.3297(+2.65\%) & 0.3207(-1.11\%) & 0.3189(+3.27\%) & -3.99\% \\
 & \textbf{0.3/0.7} & \textbf{0.616(-12.87\%)} & \textbf{0.561(-16.39\%)} & \textbf{0.64(+1.59\%)} & \textbf{0.553(+0.91\%)} & \textbf{0.0454(-3.40\%)} & \textbf{0.3296(+2.62\%)} & \textbf{0.3102(-4.35\%)} & \textbf{0.32(+3.63\%)} & \textbf{-3.53\%} \\
 & 0.4/0.6 & 0.646(-8.63\%) & 0.591(-11.92\%) & 0.653(+3.65\%) & 0.55(+0.36\%) & 0.0445(-5.32\%) & 0.322(+0.25\%) & 0.2965(-8.57\%) & 0.314(+1.68\%) & -3.56\% \\
 & 0.5/0.5 & 0.677(-4.24\%) & 0.61(-9.09\%) & 0.656(+4.13\%) & 0.569(+3.83\%) & 0.0398(-15.32\%) & 0.3048(-5.11\%) & 0.2711(-16.40\%) & 0.2996(-2.98\%) & -5.65\% \\
 & 0.6/0.4 & 0.677(-4.24\%) & 0.64(-4.62\%) & 0.656(+4.13\%) & 0.574(+4.74\%) & 0.0325(-30.85\%) & 0.2741(-14.66\%) & 0.2322(-28.40\%) & 0.273(-11.59\%) & -10.69\% \\
 & 0.7/0.3 & 0.689(-2.55\%) & 0.646(-3.73\%) & 0.651(+3.33\%) & 0.563(+2.74\%) & 0.0286(-39.15\%) & 0.2855(-11.11\%) & 0.2356(-27.35\%) & 0.2808(-9.07\%) & -10.86\% \\
 & 0.8/0.2 & 0.683(-3.39\%) & 0.628(-6.41\%) & 0.646(+2.54\%) & 0.566(+3.28\%) & 0.012(-74.47\%) & 0.2459(-23.44\%) & 0.1832(-43.51\%) & 0.2033(-34.16\%) & -22.45\% \\
 & 0.9/0.1 & 0.707(0.00\%) & 0.652(-2.83\%) & 0.64(+1.59\%) & 0.542(-1.09\%) & 0.008(-82.98\%) & 0.2291(-28.67\%) & 0.1565(-51.74\%) & 0.1892(-38.73\%) & -25.56\% \\
\hline
\multirow{9}{*}{\centering DELLA} & 0.1/0.9 & 0.573(-18.95\%) & 0.537(-19.97\%) & 0.603(-4.29\%) & 0.513(-6.39\%) & 0.0322(-31.49\%) & 0.2596(-19.18\%) & 0.226(-30.31\%) & 0.2576(-16.58\%) & -18.39\% \\
 & 0.2/0.8 & 0.555(-21.50\%) & 0.518(-22.80\%) & 0.526(-16.51\%) & 0.466(-14.96\%) & 0.0412(-12.34\%) & 0.3021(-5.95\%) & 0.2809(-13.38\%) & 0.2939(-4.83\%) & -14.03\% \\
 & \textbf{0.3/0.7} & \textbf{0.555(-21.50\%)} & \textbf{0.506(-24.59\%)} & \textbf{0.563(-10.63\%)} & \textbf{0.489(-10.77\%)} & \textbf{0.0403(-14.26\%)} & \textbf{0.2978(-7.29\%)} & \textbf{0.2743(-15.42\%)} & \textbf{0.2909(-5.80\%)} & \textbf{-13.78\%} \\
 & 0.4/0.6 & 0.567(-19.80\%) & 0.518(-22.80\%) & 0.598(-5.08\%) & 0.513(-6.39\%) & 0.0379(-19.36\%) & 0.282(-12.20\%) & 0.2538(-21.74\%) & 0.2777(-10.07\%) & -14.68\% \\
 & 0.5/0.5 & 0.616(-12.87\%) & 0.567(-15.50\%) & 0.606(-3.81\%) & 0.519(-5.29\%) & 0.0325(-30.85\%) & 0.2599(-19.08\%) & 0.2263(-30.22\%) & 0.2578(-16.52\%) & -16.77\% \\
 & 0.6/0.4 & 0.622(-12.02\%) & 0.579(-13.71\%) & 0.64(+1.59\%) & 0.542(-1.09\%) & 0.0301(-35.96\%) & 0.2455(-23.57\%) & 0.2079(-35.89\%) & 0.2445(-20.82\%) & -17.69\% \\
 & 0.7/0.3 & 0.628(-11.17\%) & 0.579(-13.71\%) & 0.638(+1.27\%) & 0.545(-0.55\%) & 0.0294(-37.45\%) & 0.2453(-23.63\%) & 0.2038(-37.16\%) & 0.2432(-21.24\%) & -17.96\% \\
 & 0.8/0.2 & 0.64(-9.48\%) & 0.604(-9.99\%) & 0.646(+2.54\%) & 0.563(+2.74\%) & 0.0282(-40.00\%) & 0.2408(-25.03\%) & 0.1977(-39.04\%) & 0.2377(-23.02\%) & -17.66\% \\
 & 0.9/0.1 & 0.665(-5.94\%) & 0.628(-6.41\%) & 0.664(+5.40\%) & 0.561(+2.37\%) & 0.0282(-40.00\%) & 0.2478(-22.85\%) & 0.2035(-37.25\%) & 0.2439(-21.02\%) & -15.71\% \\
\hline

\end{tabular}}
\caption{DeepSeek Coder 1.3B ablation study results showing performance across different merging strategies (Linear, TIES, DARE, DELLA) with varying weight ratios. Each row represents a different CodeGen/CodeSum weight combination, with performance metrics for code generation tasks (HumanEval, HumanEval+, MBPP, MBPP+) and code summarization tasks (CodeXGLUE). Percentage changes are relative to the SFT baseline. The rightmost column shows the average percentage change across all metrics.}
\label{tab:ablation_results_dsc13b}
\end{table*}

\begin{table*}[t!] 
\centering
\resizebox{\textwidth}{!}{%
\begin{tabular}{l|c|cccc|cccc|c}
\hline
Task & \multirow{3}{*}{\centering Weight} & \multicolumn{4}{c|}{\textbf{Code Generation}} &  \multicolumn{4}{c|}{\textbf{Code Summarization}} & \\
\cline{3-10} 
Dataset & & HumanEval & HumanEval+ & MBPP & MBPP+ & \multicolumn{4}{c|}{CodeXGLUE Code-to-text} & \\
Metric & & pass@1 $\uparrow$& pass@1 $\uparrow$  & pass@1 $\uparrow$  & pass@1 $\uparrow$  & BLEU-4$\uparrow$ & chrF++$\uparrow$& ROUGE-L$\uparrow$ & METEOR $\uparrow$& AVG \% wrt SFT  \\
\hline
\multicolumn{2}{c}{} & \multicolumn{8}{c}{\textit{DeepSeek Coder 7B}} &  \\
\hline
SFT-CodeGen/Sum & - & 0.793(0\%) & 0.750(0\%) & 0.775(0\%) & 0.672(0\%) & 0.0532(0\%) & 0.3277(0\%) & 0.3360(0\%) & 0.3155(0\%) & 0.00\% \\
\hline
\multirow{9}{*}{\centering Linear} & 0.1/0.9 & 0.774(-2.40\%) & 0.732(-2.40\%) & 0.783(+1.03\%) & 0.68(+1.19\%) & 0.0445(-16.35\%) & 0.3115(-4.97\%) & 0.315(-6.36\%) & 0.2959(-6.72\%) & -4.62\% \\
 & 0.2/0.8 & 0.726(-8.45\%) & 0.652(-13.07\%) & 0.77(-0.65\%) & 0.664(-1.19\%) & 0.0529(-0.56\%) & 0.3347(+2.10\%) & 0.3334(-0.89\%) & 0.324(+2.14\%) & -2.57\% \\
 & 0.3/0.7 & 0.726(-8.45\%) & 0.652(-13.07\%) & 0.772(-0.39\%) & 0.677(+0.74\%) & 0.053(-0.38\%) & 0.335(+2.20\%) & 0.331(-1.61\%) & 0.3242(+2.21\%) & -2.34\% \\
 & \textbf{0.4/0.6} & \textbf{0.738(-6.94\%)} & \textbf{0.689(-8.13\%)} & \textbf{0.775(0.00\%)} & \textbf{0.675(+0.45\%)} & \textbf{0.0512(-3.76\%)} & \textbf{0.3326(+1.46\%)} & \textbf{0.3276(-2.62\%)} & \textbf{0.3209(+1.17\%)} & \textbf{-2.30\%} \\
 & 0.5/0.5 & 0.774(-2.40\%) & 0.732(-2.40\%) & 0.783(+1.03\%) & 0.68(+1.19\%) & 0.0445(-16.35\%) & 0.3116(-4.94\%) & 0.315(-6.36\%) & 0.296(-6.68\%) & -4.61\% \\
 & 0.6/0.4 & 0.756(-4.67\%) & 0.72(-4.00\%) & 0.799(+3.10\%) & 0.701(+4.32\%) & 0.0351(-34.02\%) & 0.2748(-16.17\%) & 0.29(-13.79\%) & 0.253(-20.24\%) & -10.68\% \\
 & 0.7/0.3 & 0.756(-4.67\%) & 0.713(-4.93\%) & 0.796(+2.71\%) & 0.69(+2.68\%) & 0.0279(-47.56\%) & 0.2521(-23.09\%) & 0.2698(-19.80\%) & 0.2275(-28.28\%) & -15.37\% \\
 & 0.8/0.2 & 0.787(-0.76\%) & 0.732(-2.40\%) & 0.77(-0.65\%) & 0.672(0.00\%) & 0.0301(-43.42\%) & 0.2931(-10.59\%) & 0.2785(-17.21\%) & 0.2793(-11.95\%) & -10.87\% \\
 & 0.9/0.1 & 0.787(-0.76\%) & 0.738(-1.60\%) & 0.775(0.00\%) & 0.675(+0.45\%) & 0.0239(-55.08\%) & 0.2919(-10.95\%) & 0.2411(-28.33\%) & 0.2789(-12.07\%) & -13.54\% \\
\hline
\multirow{9}{*}{\centering TIES} & 0.1/0.9 & 0.726(-8.45\%) & 0.695(-7.33\%) & 0.788(+1.68\%) & 0.68(+1.19\%) & 0.043(-19.17\%) & 0.315(-3.90\%) & 0.3037(-9.72\%) & 0.3008(-5.17\%) & -6.36\% \\
 & 0.2/0.8 & 0.738(-6.94\%) & 0.677(-9.73\%) & 0.77(-0.65\%) & 0.667(-0.74\%) & 0.052(-2.26\%) & 0.3355(+2.35\%) & 0.3224(-4.16\%) & 0.3253(+2.55\%) & -2.45\% \\
 & \textbf{0.3/0.7} & \textbf{0.732(-7.69\%)} & \textbf{0.689(-8.13\%)} & \textbf{0.786(+1.42\%)} & \textbf{0.688(+2.38\%)} & \textbf{0.0509(-4.32\%)} & \textbf{0.3332(+1.65\%)} & \textbf{0.3178(-5.53\%)} & \textbf{0.3228(+1.77\%)} & \textbf{-2.31\%} \\
 & 0.4/0.6 & 0.762(-3.91\%) & 0.713(-4.93\%) & 0.786(+1.42\%) & 0.683(+1.64\%) & 0.0475(-10.71\%) & 0.3279(+0.03\%) & 0.3132(-6.90\%) & 0.3163(-0.28\%) & -2.96\% \\
 & 0.5/0.5 & 0.726(-8.45\%) & 0.695(-7.33\%) & 0.788(+1.68\%) & 0.68(+1.19\%) & 0.0431(-18.99\%) & 0.3151(-3.87\%) & 0.3038(-9.69\%) & 0.3009(-5.14\%) & -6.33\% \\
 & 0.6/0.4 & 0.756(-4.67\%) & 0.732(-2.40\%) & 0.796(+2.71\%) & 0.69(+2.68\%) & 0.0381(-28.38\%) & 0.2997(-8.57\%) & 0.29(-13.79\%) & 0.282(-11.10\%) & -7.94\% \\
 & 0.7/0.3 & 0.774(-2.40\%) & 0.738(-1.60\%) & 0.788(+1.68\%) & 0.683(+1.64\%) & 0.038(-28.57\%) & 0.2993(-8.69\%) & 0.29(-13.79\%) & 0.2814(-11.29\%) & -7.88\% \\
 & 0.8/0.2 & 0.756(-4.67\%) & 0.72(-4.00\%) & 0.791(+2.06\%) & 0.685(+1.93\%) & 0.0394(-25.94\%) & 0.3057(-6.74\%) & 0.2934(-12.78\%) & 0.2893(-8.80\%) & -7.37\% \\
 & 0.9/0.1 & 0.75(-5.42\%) & 0.72(-4.00\%) & 0.796(+2.71\%) & 0.69(+2.68\%) & 0.0398(-25.19\%) & 0.31(-5.43\%) & 0.2956(-12.13\%) & 0.2947(-7.09\%) & -6.73\% \\
\hline
\multirow{9}{*}{\centering DARE} & 0.1/0.9 & 0.738(-6.94\%) & 0.671(-10.53\%) & 0.743(-4.13\%) & 0.659(-1.93\%) & 0.0172(-67.67\%) & 0.2663(-18.76\%) & 0.2272(-32.46\%) & 0.2407(-24.12\%) & -20.82\% \\
 & 0.2/0.8 & 0.738(-6.94\%) & 0.677(-9.73\%) & 0.772(-0.39\%) & 0.68(+1.19\%) & 0.0208(-60.90\%) & 0.2148(-34.47\%) & 0.2501(-25.65\%) & 0.1879(-40.76\%) & -22.21\% \\
 & 0.3/0.7 & 0.756(-4.67\%) & 0.701(-6.53\%) & 0.783(+1.03\%) & 0.693(+3.13\%) & 0.024(-54.89\%) & 0.2182(-33.44\%) & 0.2584(-23.19\%) & 0.1926(-39.28\%) & -19.73\% \\
 & 0.4/0.6 & 0.787(-0.76\%) & 0.732(-2.40\%) & 0.783(+1.03\%) & 0.68(+1.19\%) & 0.0285(-46.43\%) & 0.2359(-28.04\%) & 0.2719(-19.17\%) & 0.2117(-33.26\%) & -15.98\% \\
 & 0.5/0.5 & 0.744(-6.18\%) & 0.707(-5.73\%) & 0.796(+2.71\%) & 0.685(+1.93\%) & 0.0471(-11.47\%) & 0.3229(-1.49\%) & 0.3202(-4.82\%) & 0.3096(-2.40\%) & -3.43\% \\
 & \textbf{0.6/0.4} & \textbf{0.762(-3.91\%)} & \textbf{0.734(-2.13\%)} & \textbf{0.799(+3.10\%)} & \textbf{0.701(+4.32\%)} & \textbf{0.0531(-0.19\%)} & \textbf{0.3315(+1.13\%)} & \textbf{0.3203(-4.79\%)} & \textbf{0.3196(+0.76\%)} & \textbf{-0.21\%} \\
 & 0.7/0.3 & 0.762(-3.91\%) & 0.726(-3.20\%) & 0.786(+1.42\%) & 0.693(+3.13\%) & 0.0487(-8.46\%) & 0.3312(+1.04\%) & 0.3163(-5.98\%) & 0.3200(+0.88\%) & -1.88\% \\
 & 0.8/0.2 & 0.756(-4.67\%) & 0.701(-6.53\%) & 0.78(+0.65\%) & 0.68(+1.19\%) & 0.0504(-5.26\%) & 0.3298(+0.61\%) & 0.3095(-8.00\%) & 0.3196(+0.76\%) & -2.66\% \\
 & 0.9/0.1 & 0.744(-6.18\%) & 0.701(-6.53\%) & 0.786(+1.42\%) & 0.685(+1.93\%) & 0.0501(-5.83\%) & 0.328(+0.06\%) & 0.3077(-8.53\%) & 0.3181(+0.28\%) & -2.92\% \\
\hline
\multirow{9}{*}{\centering DELLA} & 0.1/0.9 & 0.750(-5.42\%) & 0.720(-4.00\%) & 0.791(+2.06\%) & 0.696(+3.57\%) & 0.0461(-13.35\%) & 0.3230(-1.46\%) & 0.3005(-10.67\%) & 0.3118(-1.70\%) & -3.87\% \\
& \textbf{0.2/0.8} & \textbf{0.720(-9.21\%)} & \textbf{0.671(-10.53\%)} & \textbf{0.778(+0.39\%)} & \textbf{0.669(-0.45\%)} & \textbf{0.0515(-3.20\%)} & \textbf{0.3305(+0.82\%)} & \textbf{0.3117(-7.34\%)} & \textbf{0.3202(+0.95\%)} & \textbf{-3.57\%} \\
 & 0.3/0.7 & 0.732(-7.69\%) & 0.677(-9.73\%) & 0.772(-0.39\%) & 0.669(-0.45\%) & 0.0506(-4.89\%) & 0.3287(+0.27\%) & 0.3083(-8.35\%) & 0.3185(+0.41\%) & -3.85\% \\
 & 0.4/0.6 & 0.738(-6.94\%) & 0.701(-6.53\%) & 0.772(-0.39\%) & 0.672(0.00\%) & 0.0495(-6.95\%) & 0.3267(-0.34\%) & 0.3046(-9.45\%) & 0.3168(-0.13\%) & -3.84\% \\
 & 0.5/0.5 & 0.744(-6.18\%) & 0.707(-5.73\%) & 0.786(+1.42\%) & 0.693(+3.13\%) & 0.0458(-13.91\%) & 0.3234(-1.34\%) & 0.3005(-10.67\%) & 0.3124(-1.51\%) & -4.35\% \\
 & 0.6/0.4 & 0.756(-4.67\%) & 0.72(-4.00\%) & 0.788(+1.68\%) & 0.69(+2.68\%) & 0.0459(-13.72\%) & 0.324(-1.16\%) & 0.3014(-10.40\%) & 0.3132(-1.26\%) & -3.86\% \\
 & 0.7/0.3 & 0.744(-6.18\%) & 0.713(-4.93\%) & 0.788(+1.68\%) & 0.685(+1.93\%) & 0.0455(-14.47\%) & 0.3241(-1.13\%) & 0.3013(-10.43\%) & 0.3128(-1.39\%) & -4.37\% \\
 & 0.8/0.2 & 0.732(-7.69\%) & 0.701(-6.53\%) & 0.788(+1.68\%) & 0.685(+1.93\%) & 0.0459(-13.72\%) & 0.3245(-1.01\%) & 0.3013(-10.43\%) & 0.3131(-1.29\%) & -4.63\% \\
 & 0.9/0.1 & 0.732(-7.69\%) & 0.707(-5.73\%) & 0.783(+1.03\%) & 0.685(+1.93\%) & 0.0455(-14.47\%) & 0.3246(-0.98\%) & 0.3026(-10.05\%) & 0.3129(-1.36\%) & -4.66\% \\
\hline

\end{tabular}}
\caption{DeepSeek Coder 7B ablation study results showing performance across different merging strategies (Linear, TIES, DARE, DELLA) with varying weight ratios. Each row represents a different CodeGen/CodeSum weight combination, with performance metrics for code generation tasks (HumanEval, HumanEval+, MBPP, MBPP+) and code summarization tasks (CodeXGLUE). Percentage changes are relative to the SFT baseline. The rightmost column shows the average percentage change across all metrics.}
\label{tab:ablation_results_dsc7b}
\end{table*}

This comprehensive ablation study examined model merging strategies across four code language model architectures and weight ratio in the range $[0.1, 0.2, 0.3, 0.4, 0.5, 0.6, 0.7, 0.8, 0.9]$, revealing key insights about weight optimization, architectural dependencies, and multi-task performance trade-offs. All results are summarized in Tables \ref{tab:ablation_results_qwc15b}, \ref{tab:ablation_results_qwc7b}, \ref{tab:ablation_results_dsc13b}, and \ref{tab:ablation_results_dsc7b}.

\subsection{Weight Ratio and Architectural Dependencies
}
The optimal weight configurations showed clear architectural biases. Qwen2.5-Coder models consistently performed best with 0.2/0.8 weight ratios favoring code summarization across multiple merging strategies, suggesting these models benefit from emphasizing summarization tasks that provide transferable semantic understanding. DeepSeek Coder models demonstrated more balanced preferences, with the 1.3B variant optimal at 0.3/0.7 ratios and the 7B model showing strategy-dependent variations. This indicates Qwen models have structural biases toward summarization-heavy merging, while DeepSeek models achieve better task balance through uniform weight distributions.

\subsection{Scale-Dependent and Parameter Efficiency} 
Performance improvements scaled non-linearly with model size, revealing critical insights about parameter capacity in multi-task scenarios. Smaller models (1.3B-1.5B) suffered larger performance degradations (-2.75\% to -18.66\%) across all strategies, suggesting limited parameters constrain dual-task competency through post-hoc merging. Larger 7B models preserved performance much better, with DARE achieving near-baseline results (-0.21\%) on DeepSeek Coder 7B and DELLA reaching -2.07\% on Qwen2.5-Coder 7B. This scale-dependent behavior indicates successful merging requires sufficient parameter redundancy to accommodate multiple specializations without catastrophic interference.

\subsection{Strategy Characteristics and Task Asymmetries}
Each strategy exhibited distinct behavioral signatures. DELLA consistently achieved superior overall performance on larger models through adaptive parameter selection, while DARE demonstrated remarkable stability with balanced weight ratios by preserving critical parameters from both tasks. TIES showed moderate consistency with generation-heavy preferences, and Linear merging provided reliable but generally inferior baselines. Performance patterns revealed fundamental asymmetries between tasks: code generation metrics degraded gracefully under merging due to the task's structured nature, while summarization metrics showed more dramatic sensitivity to parameter configurations. The consistent success of summarization-heavy weight ratios suggests semantic understanding from summarization provides transferable benefits to generation tasks, positioning summarization as a fundamental capability for multi-task code understanding.
\section{Additional Results for Qwen2.5 coder 14B model}

To further validate our conclusion, we conducted additional experiments on the Qwen2.5-Coder 14B model, comparing the DARE merging method against data-mixture SFT. The DARE merging was performed with weight ratio 0.3/0.7 (code generation/summarization) and density 0.5/0.5 for both tasks. As shown in Table \ref{tab:add_results}, the results demonstrate that model merging can achieve competitive or even better performance with significantly reduced training costs.

\begin{table*}[t!] 
\centering
\caption{Performance of Merged vs. Fine-Tuned Models on Code Generation and Summarization. We report code generation success (pass@1 on HumanEval and MBPP) and code summarization quality (BLEU-4, chrF++, ROUGE-L, METEOR) for Qwen2.5-Coder 14B. Pretrained = zero-shot base model; SFT-CodeGen = fine-tuned on code generation; SFT-CodeSum = fine-tuned on summarization; Data-Mixture = fine-tuned on combined data; DARE = merged method with weight ratio 0.3/0.7 (code generation/summarization) and density 0.5/0.5.}
\label{tab:add_results}
\resizebox{\textwidth}{!}{
\begin{tabular}{l|cccc|cccc|c}
\hline
Task & \multicolumn{4}{c|}{\textbf{Code Generation}} &  \multicolumn{4}{c|}{\textbf{Code Summarization}} &\\ 
Dataset  & HumanEval & HumanEval+ & MBPP & MBPP+ & \multicolumn{4}{c|}{CodeXGLUE Code-to-text} &  \\
Metric & pass@1 $\uparrow$& pass@1 $\uparrow$  & pass@1 $\uparrow$  & pass@1 $\uparrow$  & BLEU-4$\uparrow$ & chrF++$\uparrow$& ROUGE-L$\uparrow$ & METEOR $\uparrow$& AVG \% wrt SFT  \\
\hline
\multicolumn{10}{c}{} \\
\multicolumn{1}{c}{} & \multicolumn{9}{c}{\textit{Qwen2.5-Coder 14B}} \\
\hline
Base Model & 0.915(-0.65\%) & 0.854(0\%) & 0.854(-5.32\%) & 0.725(-5.84\%) & 0.0198(-68.82\%) & 0.3154(-11.36\%) & 0.2656(-18.05\%) & 0.3051(-12.45\%) & -15.31\% \\
SFT-CodeGen/Sum  & 0.921(0\%) & 0.854(0\%) & 0.902(0\%) & 0.770(0\%) & 0.0635(0\%) & 0.3558(0\%) & 0.3241(0\%) & 0.3485(0\%) & 0.00\% \\
 Data-Mixture SFT  & 0.915(-0.65\%) & 0.835(-2.22\%) & 0.899(-0.33\%) & 0.751(-2.47\%) & 0.0603(-5.04\%) & 0.3501(-1.60\%) & 0.3216(-0.77\%) & 0.3400(-2.44\%) & -1.94\% \\
 DARE Merge  & \textbf{0.896(-2.71\%)} & \textbf{0.848(-0.70\%)} & \textbf{0.886(-1.77\%)} & \textbf{0.751(-2.47\%)} & \textbf{0.0626(-1.42\%)} & \textbf{0.3524(-0.96\%)} & \textbf{0.3298(+1.76\%)} & \textbf{0.3434(-1.46\%)} & \textbf{-1.15\%} \\
\hline
\end{tabular}
}
\end{table*}
\end{document}